\documentclass[10pt,twocolumn,letterpaper]{article}
\usepackage{cvpr}              
\usepackage[dvipsnames]{xcolor}
\usepackage{booktabs}
\usepackage{graphicx}
\usepackage{multirow}
\usepackage{tabularx}
\usepackage{pifont}
\usepackage{float}

\definecolor{cvprblue}{rgb}{0.21,0.49,0.74}
\usepackage[pagebackref,breaklinks,colorlinks,citecolor=cvprblue]{hyperref}

\title{EnvPoser: Environment-aware Realistic Human Motion Estimation from Sparse Observations with Uncertainty Modeling}

\author{Songpengcheng Xia$^1$
\quad
Yu Zhang $^{1}$
\quad
Zhuo Su$^2$ \thanks{Project Lead}
\quad
Xiaozheng Zheng$^2$
\quad
Zheng Lv$^2$
\quad
Guidong Wang$^2$
\\
Yongjie Zhang$^2$
\quad
Qi Wu$^{1,2}$
\quad
Lei Chu $^{3}$ 
\quad
Ling Pei$^{1}$   \thanks{Corresponding Author \\ This work was supported by the National Nature Science Foundation of China (NSFC) under Grant 62273229.}
\\
$^1$Shanghai Jiao Tong University \quad
$^2$ByteDance
\quad
$^3$University of Southern California
}

\begin{document}

\maketitle


\begin{abstract}
Estimating full-body motion using the tracking signals of head and hands from VR devices holds great potential for various applications. However, the sparsity and unique distribution of observations present a significant challenge, resulting in an ill-posed problem with multiple feasible solutions (i.e., hypotheses). This amplifies uncertainty and ambiguity in full-body motion estimation, especially for the lower-body joints. Therefore, we propose a new method, EnvPoser, that employs a two-stage framework to perform full-body motion estimation using sparse tracking signals and pre-scanned environment from VR devices. EnvPoser models the multi-hypothesis nature of human motion through an uncertainty-aware estimation module in the first stage. In the second stage, we refine these multi-hypothesis estimates by integrating semantic and geometric environmental constraints, ensuring that the final motion estimation aligns realistically with both the environmental context and physical interactions.
Qualitative and quantitative experiments on two public datasets demonstrate that our method achieves state-of-the-art performance, highlighting significant improvements in human motion estimation within motion-environment interaction scenarios. Project page: \url{https://xspc.github.io/EnvPoser/}.

\end{abstract}    
\section{Introduction}
\label{sec:intro}

In recent years, motion capture algorithms using sparse tracking signals have achieved significant progress across various research fields, establishing themselves as essential technologies for applications such as motion-sensing games and immersive VR/AR experiences~\cite{huang2018deep, kaufmann2021pose, yi2022physical, xia2024timestamp}.
With the rapid development of VR devices, full-body motion estimation algorithms based on head-mounted display (HMD) have shown great application potential. However, devices like PICO and Quest typically provide sparse tracking signals of the head and hands, presenting challenges in estimating the full-body motion from these limited observations.

\begin{figure}[t]
    \centering
    \includegraphics[width=0.48\textwidth]{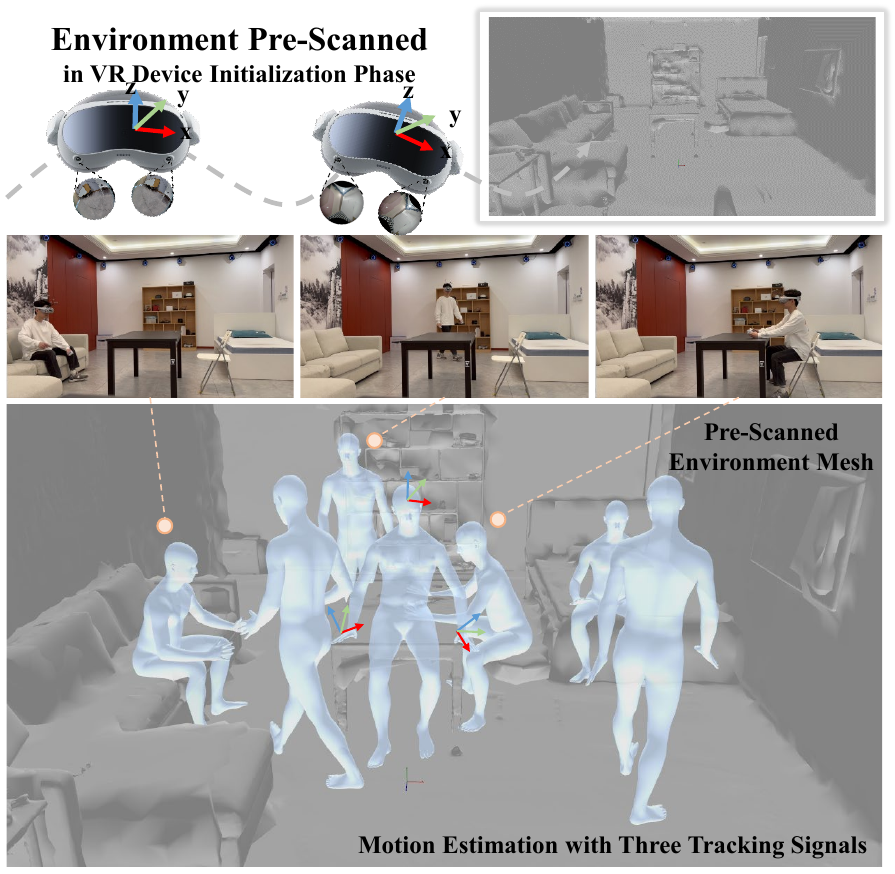} 
    \vspace{-5mm}
    \caption{
      EnvPoser can estimate the full-body motion using three tracking signals (HMD and hand controllers) and a pre-scanned environment mesh.
    }
    \label{fig:1}
    \vspace{-5mm}
    
\end{figure}

Current sparse-observation-based human motion estimation methods typically rely on large-scale motion capture datasets, such as AMASS~\cite{mahmood2019amass}, to capture full-body motion through regression~\cite{jiang2022avatarposer, zheng2023realistic} or generative models~\cite{feng2024stratified, du2023avatars, aliakbarian2023hmd}.
Alternatively, physics engines and reinforcement learning have been used~\cite{winkler2022questsim} to synthesize full-body motion, bypassing dataset dependency.
Utilizing various advanced architectures or large-scale datasets, these methods~\cite{feng2024stratified, du2023avatars, aliakbarian2023hmd, zheng2023realistic, winkler2022questsim} attempt to model the mapping from sparse tracking signals to the most probable full-body motion. However, the lack of direct joint observations results in the same sparse input often corresponding to multiple plausible motions, making it a challenge for sparse-observation-based motion estimation algorithms to converge on the most probable full-body motion. To address this one-to-many mapping ambiguity, existing methods~\cite{ponton2023sparseposer, dai2024hmd, ponton2024dragposer} attempt to add sensors to regions such as the pelvis and legs to narrow the distribution of possible motion outcomes. Although additional sensors can enhance the accuracy and robustness of full-body motion estimation, this approach may negatively impact user experience due to increased hardware requirements~\cite{huang2018deep, jiang2022transformer, yi2022physical, zhang2024dynamic, xia2021learning}.

Therefore, an alternative approach to addressing the inherent ambiguity in sparse observations is to incorporate additional environmental constraints, as shown in Fig.~\ref{fig:1}. Human motion is highly correlated with the surrounding environment, which contains rich context that can provide valuable cues for motion estimation. However, effectively utilizing environmental constraints presents its own challenges, as existing methods~\cite{tang2024unified, zheng2023realistic} often simplify human-environment interactions to basic foot-to-ground contact, overlooking the complexities of interactions. In reality, individuals frequently engage in more complex interactions with their surroundings.

Our key insight is that joint uncertainty estimation can explicitly model the multi-hypothesis nature of full-body motion estimation, and by incorporating environmental information, we can reduce estimation uncertainty, guiding the motion estimation to converge toward the most plausible outcome that aligns with both the environment and sparse input observations. 
Therefore, we design a two-stage deep learning framework to handle the one-to-many mapping ambiguity challenge. We first explicitly model the multi-hypothesis human motion by the joint uncertainty estimation and its sampling strategy. Then, by integrating environmental information, we guide the uncertainty-based multi-hypothesis motion estimations to converge on a full-body motion that best aligns with both the environment and sparse observations. 
To address the challenge of effectively utilizing environmental information, we introduce semantic constraints for non-contact motions, providing contextual cues for realistic estimation, and geometric constraints for contact scenarios, ensuring consistency and preventing collisions.

The key contributions of this paper are as follows:

\begin{itemize}
    \item We propose EnvPoser, a novel framework that significantly advances full-body motion estimation from sparse tracking signals by incorporating pre-scanned environmental context. This framework addresses the complexities of estimating human motion from sparse observations, ensuring robustness across diverse interactive scenarios.

     \item This work presents a two-stage framework for human motion estimation: an uncertainty-aware module provides initial estimates that account for multi-hypothesis motion, while an environment-aware refinement module integrates semantic and geometric constraints to ensure realistic outcomes in diverse scenarios, setting a new benchmark in motion estimation with environmental interactions.

    
    
    \item Comprehensive quantitative and qualitative experiments demonstrate that our proposed method achieves state-of-the-art performance across two public datasets and excels in scenarios with frequent environmental interactions.

\end{itemize}
\section{Related Work}
\label{sec:Related Work}

\subsection{Motion estimation with sparse observations}
Current Wearable-based human motion estimation approaches mainly adopt two configurations: one uses multiple inertial measurement units (IMUs) placed on body extremities~\cite{jiang2022transformer, zhang2024dynamic, huang2018deep, yi2021transpose, wu2024accurate, van2024diffusionposer}, while the other relies on an HMD with two controllers~\cite{jiang2022avatarposer, zheng2023realistic, dai2024hmd, jiang2025egoposer, aliakbarian2023hmd, fan2024emhi}.


While methods utilizing six IMUs have shown impressive results, HMD-based techniques leveraging sparse three-point observations show significant potential for practical applications. For instance, AvatarPoser \cite{jiang2022avatarposer} employs a transformer architecture to achieve accurate full-body motion. While Zheng et al. \cite{zheng2023realistic} developed a two-stage framework to capture joint dependencies, improving motion accuracy and fluidity. Some studies \cite{dittadi2021full, aliakbarian2022flag, feng2024stratified} address the ill-posed problem of sparse-to-dense mapping using generative models. Dittadi et al. \cite{dittadi2021full} and Aliakbarian et al. \cite{aliakbarian2022flag} proposed using Variational Autoencoders (VAE) and normalizing flows, respectively, to estimate full-body poses from three tracking points. Additionally, the diffusion model \cite{du2023avatars, feng2024stratified, tang2024unified} has also shown notable performance in motion estimation with sparse data.
Despite these advances, most methods rely solely on sparse tracking signals, leading to high uncertainty and multiple potential poses for the same input. Therefore, this paper introduces joint uncertainty estimation and constrains physically implausible motions using environmental information.

\subsection{Pose regression with uncertainty estimation}
The sparse nature of observations in 3D human pose estimation introduces significant uncertainty for unobserved joints, as identical inputs can correspond to multiple pose hypotheses~\cite{feng2024stratified, zheng2023realistic, zhang2024dynamic}.  This challenge is particularly evident in image-based pose estimation \cite{li2022mhformer, li2019generating, xu2024scorehypo, gong2023diffpose, chen2023mhentropy}, where monocular methods face depth ambiguity. For instance, Wehrbein et al. \cite{wehrbein2021probabilistic} modeled the posterior distribution of 3D pose hypotheses using normalizing flows, while MHFormer \cite{li2022mhformer} employed a multi-hypothesis transformer to capture diverse pose hypotheses through a one-to-many mapping. Recently, Shan et al. \cite{shan2023diffusion} proposed diffusion-based methods to aggregate multi-hypothesis predictions with joint-wise re-projection.

In our paper, we aim to use uncertainty estimation to model the multi-hypothesis nature of motion from sparse observations. Uncertainty in deep learning can be categorized into two types \cite{kendall2017uncertainties}: aleatoric (data-related) and epistemic (model-related). Li et al. \cite{li2023pose} refined 3D poses by incorporating joint uncertainties through uncertainty guided-sampling and uncertainty-guided self-attention, while Zhang et al. \cite{zhang2023body} introduced a probabilistic framework that encodes aleatoric uncertainty using a robust negative log-likelihood loss, with epistemic uncertainty guiding model refinement. In wearable motion estimation, Yang et al. \cite{yang2024spatial} leveraged uncertainty to weight IMU-derived features, enhancing pose corrections with additional text annotations.


\subsection{Scene-aware motion estimation and generation}
Recent advancements in human motion estimation and generation have expanded from solely analyzing body motion to integrating interactions with the surrounding environment, especially vision-based approaches~\cite{shen2023learning, zhang2023probabilistic, jiang2024scaling, yi2023mime, huang2023diffusion, huang2022capturing}. For instance, PROX~\cite{hassan2019resolving} incorporated scene constraints into monocular human pose estimation to address depth ambiguity, while Shen et al. \cite{shen2023learning} used sparse 3D CNNs to estimate absolute positions and dense scene contacts, refining human mesh recovery via cross-attention with 3D scene cues. EgoHMR \cite{zhang2023probabilistic} leveraged ego-camera images and diffusion models within pre-scanned environments to recover full-body motion. Similarly, S2Fusion \cite{tang2024unified} utilized sparse observations by learning human motion priors on synthetic datasets and conditioned motion generation on scene encoding and VR inputs. However, S2Fusion primarily focuses on lower body and environment interactions. QuestEnvsim \cite{lee2023questenvsim} employed reinforcement learning to generate realistic poses in highly constrained environments using HMDs.
Additionally, recent studies further emphasize the integration of scene context and historical motion to produce plausible human motion~\cite{xuan2023narrator, lou2024multimodal, yang2024smgdiff}. 


These scene-aware methods underscore the importance of environmental context for enhancing motion estimation. Our approach builds on these foundations by comprehensively leveraging human-environment interactions, addressing both contact and non-contact scenarios to improve motion estimation with sparse tracking signals.

\section{Method}
\subsection{Problem Statement}
Our goal is to estimate the full-body motion sequence $\boldsymbol{\theta}=\{\theta_t\}^T_{t=1} \in \mathcal{R}^{T \times 132}$ (the first 22 joints of SMPL~\cite{loper2015smpl} model with the 6D representation of rotations) using the sparse tracking signals $\boldsymbol{X}=\{{\boldsymbol x}_t\}^T_{t=1} \in \mathcal{R}^{T \times N_c}$ from the HMD and hand controllers, along with a pre-scanned 3D environment point cloud $\boldsymbol{S}$, where $T$ denotes the length of input, $N_c$ = 36 means the channel of input. Each sparse tracking signal at time step $t$ consists of the head and hand positions $\boldsymbol{p}_t$, rotations $\boldsymbol{R}_t$, and linear velocities $\boldsymbol{v}_t$, represented as $\boldsymbol{x}_t = [\boldsymbol{p}_t, \boldsymbol{R}_t, \boldsymbol{v}_t]$.
Moreover, we utilize a cropped 3D environment point cloud $\boldsymbol{V_S} \in \mathcal{R}^{N_S \times 3}$, where $N_S$, the number of sampled points, is set to 1000.
With the human translation, this point cloud refines pose estimation by incorporating environmental context, leveraging both contact and non-contact interactions between the body and the environment to reduce the uncertainty inherent in sparse observations.

\begin{figure*}[h]
	\centering
	\includegraphics[width=17.5cm]{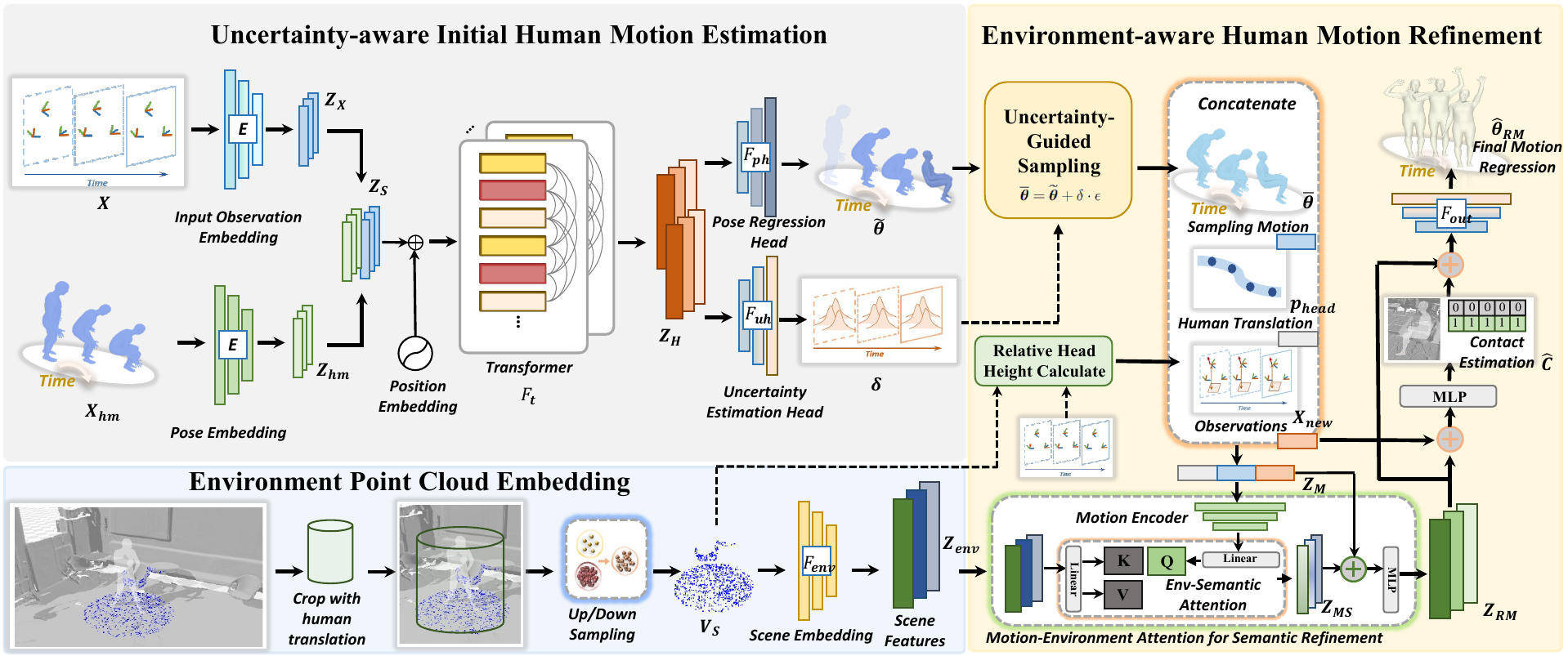}
    \vspace{-5mm}
	\caption{{Overview of EnvPoser: A Two-Stage Motion Estimation Model. Stage I involves training the uncertainty-aware initial estimation module on the AMASS dataset to produce initial motion estimates with uncertainty quantification. Stage II refines these estimates by training on motion-environment datasets, incorporating semantic and geometric environmental constraints.}}
	\label{overview}
	\vspace{-5mm}
\end{figure*}

As shown in Fig.~\ref{overview}, our approach is a two-stage model comprising two core technical modules: 1) an uncertainty-aware initial motion estimation module; and 2) an environment-aware motion refinement module that leverages semantic and geometric information from the environment. We first train the uncertainty-aware initial motion estimation module (Stage I) on the AMASS dataset, with a focus on explicitly capturing the multi-hypothesis nature of motion estimation through uncertainty quantification. Subsequently, we jointly train both modules (Stage II) on the motion-environment interaction datasets, applying environmental semantic and geometric constraints to refine the multi-hypothesis estimates from Stage I into the most plausible motion estimation.


\subsection{Uncertainty-aware Initial Human Motion Estimation Module}

To obtain the joint uncertainty quantification and explicitly model the multi-hypothesis nature of motion estimation, we propose an uncertainty-aware initial human motion estimation module. This module is designed with three primary objectives: (1) to integrate historical motion states and sparse observations for the current initial motion estimation; (2) to effectively extract features from sparse signals; and (3) to quantify uncertainty in motion estimation, explicitly modeling the multi-hypothesis nature of motion under identical input conditions. Accordingly, this module comprises three sub-modules: sparse input signal and historical motion embedding, transformer-based motion feature extraction, and human motion and joint uncertainty regression.

\textbf{Sparse Input Signal and Historical Human Motion Embedding.}
To integrate historical motion states and sparse observations for the current initial motion estimation, we adopt an auto-regressive structure in the uncertainty-aware initial human motion estimation module.
Our model takes sparse observations $\boldsymbol{X} \in \mathcal{R}^{T \times 36}$ and historical motion $\boldsymbol{X}_{hm} \in \mathcal{R}^{T \times 132}$ as inputs, embedding these inputs through linear layers to obtain the observation representation $\boldsymbol{Z}_X$ and motion historical state representation $\boldsymbol{Z}_{hm}$. 
Then, we concatenate the sparse observation embedding and historical motion embedding to obtain the shallow motion representation $\boldsymbol{Z}_S = [\boldsymbol{Z}_X, \boldsymbol{Z}_{hm}]$, which can be used for subsequent transformer-based feature extraction.

\textbf{Transformer-based Motion Feature Extraction.}
Based on the shallow motion representation $\boldsymbol{Z}_S$, we employ a transformer-based feature extraction module~\cite{jiang2022transformer} for motion features learning. To retain sequence order information, temporal positional encoding is applied to the shallow motion representation~\cite{vaswani2017attention}. Using multi-head self-attention and a position-wise feed-forward network within the transformer encoder $F_{t}$, this module efficiently captures intricate relationships in the sparse tracking signals and historical motion, resulting in high-dimensional motion features $\boldsymbol{Z}_{H}$ = $F_{t}(\boldsymbol{Z}_S)$. 
While the transformer architecture is effective for extracting motion features, the sparsity of the input data—limited to tracking signals from the head and hands—introduces high uncertainty in estimating unobserved joints, particularly in the lower body. This uncertainty arises from the lack of direct observations for these joints, making their estimation inherently ambiguous. To address this challenge, we incorporate a mechanism to quantify the uncertainty in joint estimations.

\textbf{Human Motion and Joint Uncertainty Regression.}
With the high-dimensional motion features $\boldsymbol{Z}_{H}$, we employ a pose regression network $F_{ph}(\cdot)$ to predict the human motion $\widetilde{\boldsymbol{\theta}} = F_{ph}(\boldsymbol{Z}_{H})$. Given the sparse nature of the input signals, capturing joint uncertainty is crucial in mapping the limited observations to a full-body motion estimation. To achieve this, we utilize a heteroscedastic neural network~\cite{kendall2017uncertainties, li2023pose} to estimate the joint uncertainty.
Specifically, an uncertainty regression head $F_{uh}(\cdot)$ is employed on the high-dimensional motion feature to estimate the joint uncertainty $\boldsymbol{\delta} \in \mathcal{R}^{T \times 132}$ of the predicted motion. 

Using the predicted motion and its associated uncertainty, we resample the full-body motion $\overline{\boldsymbol{\theta}}$ around the predicted mean motion $\widetilde{\boldsymbol{\theta}}$ according to a Gaussian distribution $\mathcal{N}(\widetilde{\boldsymbol{\theta}}, \boldsymbol{\delta})$ with the uncertainty $\delta$. This resampling technique introduces a multi-hypothesis representation, characterizing the inherent ambiguity in the unobserved joints. To ensure differentiability throughout the model training process, we apply the reparameterization trick on this sampling operation, drawing a parameter $\epsilon$ from a standard Gaussian distribution $\mathcal{N}(0,1)$, represented by $\overline{\boldsymbol{\theta}}$ = $\widetilde{\boldsymbol{\theta}} + \boldsymbol{\delta} \cdot \epsilon$.

Through the uncertainty regression head, we can obtain the joint uncertainty estimation for predicted human motion, enabling a multi-hypothesis motion representation with the sampling strategy.
We pre-train the first stage of our method on the large synthetic dataset AMASS~\cite{mahmood2019amass}. Based on the predicted mean human motion $\widetilde{\boldsymbol{\theta}}$ and uncertainty $\boldsymbol{\delta}$, we train the first stage module with the following loss function:
\begin{equation}
\begin{aligned}
& \label{motionloss}
   L_{M} = \|\widetilde{\boldsymbol{\theta}} - \boldsymbol{\theta}\|_2,
\end{aligned}%
\end{equation}
where $\boldsymbol{\theta}$ represents the ground-truth of human motion. Additionally, to ensure accurate uncertainty prediction, we also incorporate an uncertainty estimation loss as \cite{kendall2017uncertainties} stated:
\begin{equation}
\begin{aligned}
& \label{deltaloss}
   L_{\delta} = \| \frac{\widetilde{\boldsymbol{\theta}} - \boldsymbol{\theta}}{\boldsymbol{\delta}} \|_2 + \log(\| \boldsymbol{\delta} \|_2).
\end{aligned}%
\end{equation}

Thus, based on the uncertainty-aware initial human motion module, the objective function for Stage I is defined as follows:
\begin{equation}
\begin{aligned}
& \label{Fir-all}
L_{S-I} = \lambda_M L_M + \lambda_{\delta} L_{\delta},
\end{aligned}%
\end{equation}
where the $\lambda_M$ and $\lambda_{\delta}$ are the hyper-parameters for Stage I training.

Training Stage I on the AMASS dataset, along with the design for motion uncertainty quantification, allows our network to effectively capture the multi-hypothesis nature of motion estimation arising from sparse observations. To address the inherent one-to-many ambiguity, we introduce environmental information as a refinement stage. By incorporating both semantic and geometric environmental information, our method guides the multi-hypothesis motion estimates to converge toward the solution that best aligns with the sparse observations and environmental constraints.

\subsection{Environment-aware Human Motion Refinement with Semantic and Geometry Constraint}

Based on the initial motion estimation and its uncertainty produced by the first-stage uncertainty-aware module, we further refine the multi-hypothesis human motion estimates using the pre-scanned environment point clouds collected by real-world devices. 
To effectively utilize the environmental information, our design is guided by a key insight: human motion within an environment involves both object interactions (\textit{contact scenarios}) and collision avoidance (\textit{non-contact scenarios}).
In this module, the environment point clouds are processed through cropping, sampling, and encoding. We then apply constraints based on the environment's semantic and geometric properties to ensure that the full-body motion estimation realistically aligns with the surrounding context.

\textbf{Environmental Point Cloud Embedding.}
To enhance computational efficiency and focus on relevant interaction areas, we first clip the environmental point clouds based on the human’s global position. A circular sampling approach is used to minimize the impact of body orientation, selecting a 1-meter radius around the body and uniformly distributing $N_S$ points within this area, denoted as $\boldsymbol{V_S} \in \mathcal{R}^{N_S \times 3}$. The cropped environmental point cloud $\boldsymbol{V_S}$  is then processed by the environment encoder $F_{env}(\cdot)$ to obtain the environmental point cloud embedding $\boldsymbol{Z}_{env}$, represented by:
\begin{equation}
\begin{aligned}
& \label{sceneem}
   \boldsymbol{Z}_{env} = F_{env}(\boldsymbol{V_S}),
\end{aligned}%
\end{equation}
where the environment encoder $F_{env}(\cdot)$ is implemented using the vanilla PointNet++ \cite{qi2017pointnet++}. The extracted environmental embedding would be used in the environment semantic-aware motion refinement.

\textbf{Environment Semantic-Aware Constraint for Motion Refinement.}
Capturing semantic information from the environmental point cloud provides crucial guidance for motion generation~\cite{lou2024multimodal, xuan2023narrator, yi2023mime}, as human motion is highly correlated with environmental context. Therefore, leveraging this motion-environment correlation as a valuable supplement has the potential to enhance the sparse-observation-based human motion estimation.
Inspired by \cite{lou2024multimodal, shen2023learning}, we employ a cross-attention module between the initial motion estimates and the cropped environmental point clouds to refine the human motion estimation. This module comprises an environment-embedding network, a motion information embedding network and an environment-motion cross-attention network.

The motion information embedding network $F_{motion}(\cdot)$ takes the sampled motion estimation $\widetilde{\boldsymbol{\theta}}$, head translation $\boldsymbol{p}^{head}$ and the extended sparse observations $\boldsymbol{X}_{new} = \{\boldsymbol{X}, \boldsymbol{h}, \boldsymbol{\theta}^{up} \} \in \mathcal{R}^{T \times 40}$ as input. The extended sparse observations $\boldsymbol{X}_{new}$ include the relative head height $\boldsymbol{h}$ and head up-vector $\boldsymbol{\theta}^{up}$, obtained via a relative head height calculation module. Following \cite{lee2024mocap, jiang2022transformer}, this module calculates terrain height by using the height of the feet when stationary and nearby environmental points, and then determines $\boldsymbol{h}$ as the difference between the absolute head height from the HMD and the estimated terrain height.
These inputs are concatenated and processed through linear layers to produce the motion information embedding $\boldsymbol{Z}_M = F_{motion}(concat(\overline{\boldsymbol{\theta}}, \boldsymbol{p}^{head}, \boldsymbol{X}_{new}))$.
 
For the environment-motion cross-attention network, the Key $\boldsymbol{K}$ and Value $\boldsymbol{V}$ vectors can be obtained by multiplying the weight matrix $\boldsymbol{W}_k, \boldsymbol{W}_v$ with the environmental embedding $\boldsymbol{Z}_{env}$, while the motion information embedding $\boldsymbol{Z}_M$ multiplies the weight matrices $\boldsymbol{W}_q$ to get the Query $\boldsymbol{Q}$ vectors. 
Inspired by \cite{liu2022swin, lou2024multimodal}, we incorporate a spatial salience $\boldsymbol{s}_{spatial}$ = $MLP(\boldsymbol{V_S})$ that expects our cross-attention model to be able to perceive the distance between environmental points clouds and human bodies.
Using the spatial salience $\boldsymbol{s}_{spatial}$ in conjunction with attention scores $Attn(Q, K)$, we derive a motion-environment representation $\boldsymbol{Z}_{ME}$, which could be represented by 
\begin{equation}
\begin{aligned}
& \label{ME}
   \boldsymbol{Z}_{ME} = (Attn(\boldsymbol{Q}, \boldsymbol{K}) + \boldsymbol{s}_{spatial}) \cdot \boldsymbol{V}.
\end{aligned}%
\end{equation}

Then, by concatenating the motion-environment representation $\boldsymbol{Z}_{ME}$ with the motion information embedding $\boldsymbol{Z}_M$ and passing them through an MLP network, we obtain the environment-refined motion representation $\boldsymbol{Z}_{RM}$ = $MLP(concat(\boldsymbol{Z}_{ME}, \boldsymbol{Z}_M))$.

Using the environment-refined motion representation $\boldsymbol{Z_{RM}}$, which integrates environmental semantics and spatial priors between the human and environment, the model can then regress the probability of human-environment contact and generate the final motion estimation.

\textbf{Motion Regression with Contact Estimation.}
With the environment-refined motion representation $\boldsymbol{Z}_{RM}$, we estimate the contact probability $\boldsymbol{C}$ = $\{ \boldsymbol{c} \}_{t=1}^T \in \mathcal{R}^{T \times 22}$ and regress the final human motion $\hat{\boldsymbol{\theta}}_{RM}$. We take the environment-refined motion representation $\boldsymbol{Z}_{RM}$ and the extended sparse observations $\boldsymbol{X}_{new}$ as input, and apply two-layers MLP layers on the concatenated input to get the contact probability prediction $\hat{\boldsymbol{C}}$ = $MLP(concat(\boldsymbol{X}_{new}, \boldsymbol{Z_{RM}}))$. 

Contact probability prediction is typically treated as a binary classification problem, where Binary Cross Entropy (BCE) loss is used for training:
\begin{equation}
\begin{aligned}
& \label{Lconcact}
  L_{contact} = BCELoss(\hat{\boldsymbol{C}}, \boldsymbol{C}).
\end{aligned}%
\end{equation}
Finally, the contact probability $\hat{\boldsymbol{C}}$ and environment-refined motion representation $\boldsymbol{Z}_{RM}$ are concatenated to regress the final motion with a motion decoder network $F_{out}(\cdot)$:
\begin{equation}
\begin{aligned}
& \label{finalpose}
  \hat{\boldsymbol{\theta}}_{RM} = F_{out}(concat(\boldsymbol{Z_{RM}}, \hat{\boldsymbol{C}})),
\end{aligned}%
\end{equation}
where $F_{out}(\cdot)$ consists of two linear layers with a ReLU activation function. We then obtain the final motion loss:
\begin{equation}
\begin{aligned}
& \label{Lfm}
  L_{M'} = \| \hat{\boldsymbol{\theta}}_{RM} -  \boldsymbol{\theta} \|_2.
\end{aligned}%
\end{equation}

\textbf{Scene Geometry-Aware Constraint for Motion Refinement.}
To complement semantic-level constraints, we leverage the geometric structure of the environment to enhance motion estimation accuracy, especially in human-environment interactions (contact scenarios). A straightforward approach is to detect collisions between the estimated human motion and the environment point cloud, thereby refining unrealistic motions.
While existing methods~\cite{hassan2019resolving, zhang2021learning} often convert 3D environments into Signed Distance Fields (SDF) for collision detection, this approach can be computationally intensive. Inspired by recent advancements~\cite{zhang2023probabilistic, mihajlovic2022coap}, we instead employ the COAP model \cite{mihajlovic2022coap} to efficiently compute collisions between the human body and environmental points. With this novel model and the environment point clouds, the COAP-based collision loss~\cite{zhang2023probabilistic} can be formulated as:
\begin{equation}
\begin{aligned}
& \label{coapcollision}
   L_{coap} = \frac{1}{N_S} \sum_{i=1}^{N_S} \sigma(f_{\Theta}(V_{S_i}|\mathcal{G})) \mathbb{I}_{f_{\Theta}(V_{S_i}|\mathcal{G}) > 0} ,
\end{aligned}%
\end{equation}
where $\boldsymbol{V_{S}}$ denotes the cropped environment point clouds, with $N_S$ representing the number of points. $\Theta$ represents the transformation parameters of the human body, encompassing both the motion pose $\boldsymbol{\theta}$ and shape $\boldsymbol{\beta}$. $f_{\Theta}(V_{S_i}|\mathcal{G})$ represents the spatial relationship between the human body model and environmental points.
The function $\sigma(\cdot)$ is the $Sigmoid$ function, which maps the value of $f_{\Theta}(V_{S_i}|\mathcal{G})$ to a range between 0 and 1. 


Since most motions occur on the ground, we remove ground points from the scene to ensure that the COAP-based collision loss specifically targets interactions with surrounding objects. To account for interactions with the ground, we introduce additional constraints, including foot contact, foot height, and ground penetration losses~\cite{zheng2023realistic}:
\begin{equation}
\begin{aligned}
& \label{Lfeet}
  L_{fc} = \| (\hat{\boldsymbol{P}}_{RM}^{feet} -  \boldsymbol{P}^{feet}) \cdot  \boldsymbol{C}\|_1, \\
  & L_{gfh} = \| \hat{\boldsymbol{z}}_{P_{RM}}^{feet} -  \boldsymbol{z}^{ground} \|_1, \\
  & L_{gp} = \| (\hat{\boldsymbol{z}}_{P_{RM}}^{min} -  \boldsymbol{z}^{ground}) \cdot \boldsymbol{l} \|_1, \\
\end{aligned}%
\end{equation}
where the $\hat{\boldsymbol{z}}_{P_{RM}}^{feet}$ and $\boldsymbol{z}^{ground}$ means the height of feet joint positions and the floor height, and $\boldsymbol{l}$ denotes whether the joint is lower than the ground.


\textbf{Final Loss Function.}
In addition to the first stage loss (Eq.\ref{deltaloss}, Eq.\ref{motionloss}), COAP-based collision loss (Eq.\ref{coapcollision}), contact loss (Eq.\ref{Lconcact}), foot-ground loss (Eq.\ref{Lfeet}) and final motion loss (Eq.\ref{Lfm}), we also apply some additional losses to improve model training. With the forward kinematic chain, the joint position $\boldsymbol{P}$, $\hat{\boldsymbol{P}}_{RM}$, and the global hands positions $\boldsymbol{P}^{hand}$, $\hat{\boldsymbol{P}}_{RM}^{hand}$ could be calculated by the motion pose $\boldsymbol{\theta}$ and $\hat{\boldsymbol{\theta}}$. To penalize the errors accumulating along the kinematic chain, we apply a joint position loss ($L_{posi}$) and a hand alignment loss($L_{hAL}$):
\begin{equation}
\begin{aligned}
& \label{Lglbposi}
  L_{posi} = \| \hat{\boldsymbol{P}}_{RM} -  \boldsymbol{P} \|_2,  L_{hAL} = \| \hat{\boldsymbol{P}}_{RM}^{hand} -  \boldsymbol{P}^{hand} \|_1. \\
\end{aligned}%
\end{equation}

Therefore, with the technical analysis provided above, we can formulate the second stage objective function as follows:
\begin{equation}
\begin{aligned}
& \label{Lall}
  L_{S-II} = L_{S-I} + L_{M'} + \lambda_1 L_{posi} + \lambda_2 L_{hAL} + \lambda_3 L_{fc} \\
  & + \lambda_4 L_{contact}  +  \lambda_5 L_{gfh} + \lambda_6 L_{gp} + \lambda_7 L_{coap}.
\end{aligned}%
\end{equation}
Here, ${\left\{ {\lambda_i} \right\}_{i=1,\dots,7}}$ represents the hyperparameters. Our model is first trained on the AMASS using the Eq.~\ref{Fir-all} and then fine-tuned on a motion-environment interaction dataset with Eq.~\ref{Lall}. 
The training process and selection of hyperparameters can be found in the supplementary material.

\section{Experiments}
\label{sec:experiments}
This section introduces the datasets, metrics, and competing algorithms used in our experiments. We then compare our method EnvPoser with state-of-the-art algorithms, highlighting its advantages in motion estimation. Finally, ablation studies validate the effectiveness of each module. Additional case studies and discussions are provided in the supplementary materials.
	

\begin{figure*}[!t]
\centering
	\includegraphics[width=17cm]{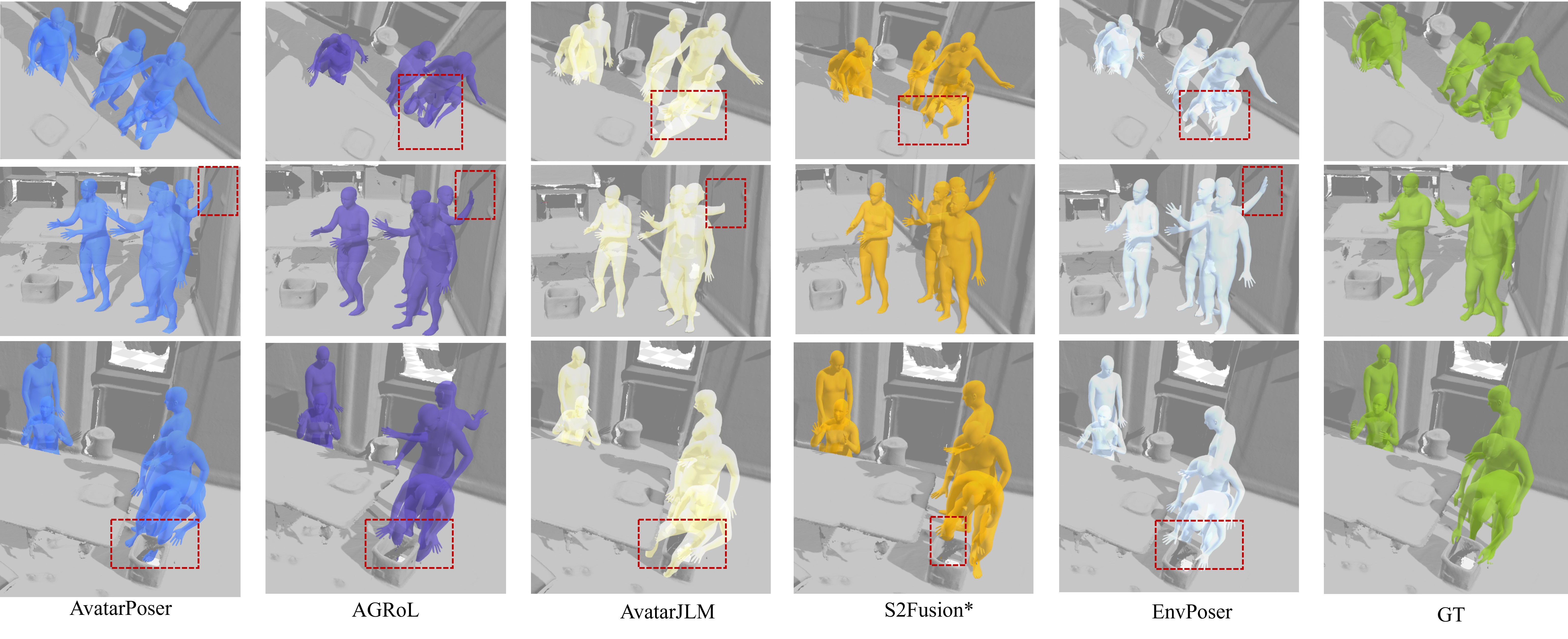}
    \vspace{-3mm}
	\caption{Visualization of motion estimation on three test sequences from EgoBody Dataset~\cite{zhang2022egobody}.}
	\label{fig:cvpr25_vis}
    \vspace{-5mm}
\end{figure*}


\textbf{Datasets. }To comprehensively evaluate the efficacy of our proposed method, we conduct experiments on two challenging public motion-scene interactions datasets: Egobody \cite{zhang2022egobody} and GIMO \cite{zheng2022gimo} datasets. Detailed information about these datasets can be found in the supplementary document.




\begin{table}
\centering

\resizebox{0.5\textwidth}{!}{
\begin{tabular}{lccccc}
\toprule
\multirow{2}{*}{} & \multicolumn{4}{c}{\textbf{EgoBody}} & \\
\cmidrule{2-5}
 & MPJRE(°) & MPJPE(mm) & MPJVE(mm/s) & Jitter \\
\midrule
AvatarPoser\cite{jiang2022avatarposer} & 6.80 & 97.8 & 237.3& 11.0\\
AGRoL\cite{du2023avatars} & 7.21 & 100.9 & 370.3 & 21.1\\
AvatarJLM\cite{zheng2023realistic} & 6.42 & 91.7 & 177.4 & 7.2 \\
S2Fusion*\cite{tang2024unified} & 6.65 & 89.2 & 219.4 & 12.5 \\
\textbf{EnvPoser} & \textbf{6.00} & \textbf{74.7} & \textbf{174.0} & \textbf{6.6}\\
\midrule
\multirow{2}{*}{} & \multicolumn{4}{c}{\textbf{GIMO}} & \\
\cmidrule{2-5}
  & MPJRE(°) & MPJPE(mm) & MPJVE(mm/s) & Jitter \\
\midrule
AvatarPoser\cite{jiang2022avatarposer} & 7.02 & 91.3 & 324.0 & 16.4\\
AGRoL\cite{du2023avatars} & 6.58 & 88.6 & 269.4 & 12.5\\
AvatarJLM\cite{zheng2023realistic} & 4.95 & 70.7 & 258.1 & 10.7 \\
S2Fusion\cite{tang2024unified} & 4.65 & 57.8 & 235.7 & 10.1 \\
\textbf{EnvPoser} & \textbf{4.38} & \textbf{57.6} & \textbf{234.6} & \textbf{8.9}\\
\bottomrule
\end{tabular}}
\vspace{-3mm}
\caption{The performance comparison with SOTAs.}
\centering
\vspace{-5mm}
\label{table:Sotas}
\end{table}

\textbf{Competing methods. }We compare the motion estimation performance of the proposed model with the state-of-the-art competing methods, including AvatarPoser~\cite{jiang2022avatarposer}, AGRoL~\cite{du2023avatars}, AvatarJLM~\cite{zheng2023realistic}, and S2Fusion~\cite{tang2024unified} on two public datasets. We re-train all competing methods on the EgoBody and GIMO datasets until convergence. For the results on the GIMO dataset, we reused the values reported in previous studies \cite{tang2024unified}. For the EgoBody dataset, we tested our trained model under real-time settings and presented the results as follows. It is worth noting that S2Fusion~\cite{tang2024unified} is not fully open-source; therefore, we reproduced S2Fusion based on the details provided in its paper and partial open-source code, marking it as S2Fusion* in this experiment.

\textbf{Metrics.}  Following~\cite{tang2024unified, zheng2023realistic}, we use these metrics for evaluation:
\textit{1) MPJRE [°]}: The mean joint rotational error across all body joints, measured in degrees; \textit{2) MPJPE [cm]}: The mean positional error per joint, measured in centimeters. \textit{3) MPJVE [cm/s]}: Represents the average error in joint velocities, expressed in centimeters per second; \textit{4) Jitter }: Measures motion smoothness by calculating the average jerk (rate of change of acceleration) across all joints, with lower values indicating smoother motion.

\subsection{Comparison with the State-of-the-Art}
\textbf{Quantitative Experimental Analysis. }
We compare our proposed method with the competing methods on the two public motion-environment interaction datasets. Among these, AvatarPoser~\cite{jiang2022avatarposer}, AGRoL~\cite{du2023avatars}, and AvatarJLM~\cite{zheng2023realistic} are the SOTAs that rely solely on sparse tracking signals. As shown in Tab.~\ref{table:Sotas}, our model, which incorporates pre-scanned environmental point cloud information, effectively constrains the uncertainty in motion reconstruction caused by sparse tracking signals. Compared to AvatarJLM~\cite{zheng2023realistic}, which does not utilize environmental information, our method achieves significant improvements on two key metrics—MPJRE and MPJPE—by 0.42 (6.5\%)/17.0 (18.5\%) and 0.57 (11.5\%)/13.1 (18.5\%) on the EgoBody and GIMO datasets, respectively. These results underscore the increased accuracy in motion estimation achieved through the integration of environmental information. 

\begin{figure}[t]
	\centering
	\includegraphics[width=8cm]{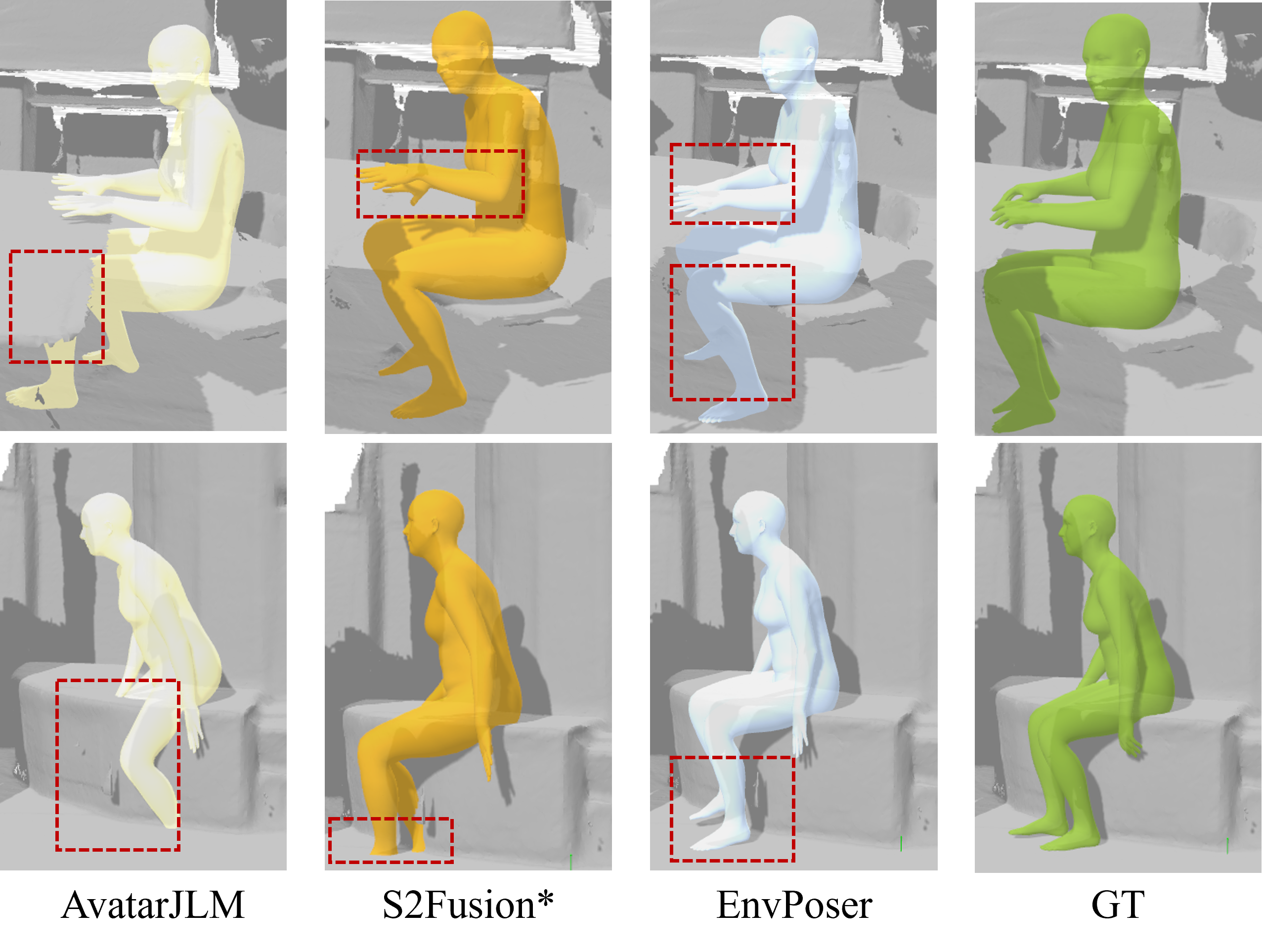}
    \vspace{-3mm}
	\caption{{Qualitative Comparison of Interaction Details.}}
	\label{vis2}
	\vspace{-5mm}
\end{figure}

Additionally, we benchmark our model against the latest algorithm S2Fusion~\cite{tang2024unified}, which introduces environmental information as a constraint for human motion estimation. On the GIMO dataset, our method reduces MPJRE errors by 5.8\% compared to S2Fusion~\cite{tang2024unified}, and it outperforms S2Fusion across all evaluation metrics, demonstrating the superior effectiveness of our approach in leveraging environmental information. Moreover, the motion estimation performance of our method also outperforms S2Fusion* on the EgoBody dataset, which reduces MPJRE and MPJPE errors by 9.8\% and 16.3\%. Furthermore, our method shows a marked improvement in motion smoothness, with an 11.9\% reduction in Jitter over S2Fusion and a 16.8\% improvement over AvatarJLM on the GIMO dataset. 

\begin{table}
\resizebox{0.45\textwidth}{!}{
\begin{tabular}{l*{4}{c}}
\toprule
& \multicolumn{2}{c}{\textbf{EgoBody}} & \multicolumn{2}{c}{\textbf{GIMO}}  \\
\cmidrule(lr){2-3} \cmidrule(lr){4-5} 
& MPJRE & MPJPE & MPJRE & MPJPE \\
\midrule
\ding{172}Baseline & 7.45 & 87.1 & 5.72 & 74.6 \\
\ding{173}w/o env-semantic & 6.04 & 78.0 & 4.55 & 62.5 \\
\ding{174}w/o env-geometry & 6.03 & 77.5 & 4.63 & 63.2 \\  
EnvPoser & \textbf{6.00} &\textbf{74.7} & \textbf{4.38} & \textbf{57.6} \\
\bottomrule
\end{tabular}
}
\vspace{-2mm}
\caption{The ablation study on environment refinement module.}
\label{table:ablation}
\vspace{-5mm}
\end{table}

\textbf{Qualitative Experimental Analysis.}
Fig.~\ref{fig:cvpr25_vis} showcases the motion estimation results on the EgoBody dataset, where we present three representative sequences and visualize the human motion estimation results at key moments. 
For state-of-the-art algorithms that do not incorporate environmental information, the visualization results in Fig.~\ref{fig:cvpr25_vis} show multiple instances of environment penetration during interactions. For example, AvatarJLM exhibits errors such as hand intersections with walls (second row) and leg penetrations through boxes (third row). Furthermore, in scenarios without chairs, squatting motions are misinterpreted as sitting (first row) in AvatarJLM due to the absence of environment understanding. In contrast, EnvPoser effectively leverages environmental information to reduce penetration errors and produce more realistic motion estimates.

In more challenging interaction scenarios, EnvPoser shows certain advantages over S2Fusion, which also incorporates environmental information. Notably, S2Fusion fails to prevent interpenetration between the lower body and surrounding objects (third row in Fig.~\ref{fig:cvpr25_vis}). Moreover, Fig.~\ref{vis2} provides a comparison between EnvPoser and S2Fusion, highlighting the advantages of our approach in handling upper-body interactions with objects and complex lower-body motions through the integration of uncertainty modeling and environmental semantic/geometric constraints. Additionally, compared to AvatarJLM, both EnvPoser and S2Fusion demonstrate the effectiveness of incorporating environmental information to improve human motion estimation.

\subsection{Ablation Study}


\textbf{Effectiveness of the environment-aware refinement module.}
We conducted ablation studies comparing our method with three variants: \ding{172} Baseline: an auto-regressive transformer-based network with joint uncertainty estimation; \ding{173} w/o env-semantic: incorporating only geometric constraints; \ding{174} w/o env-geometry: incorporating only semantic constraints.
Tab.~\ref{table:ablation} presents the motion estimation performance for each variant. Compared to variant \ding{172}, our method achieves substantial improvements by integrating both semantic and geometric environmental information. Further analysis shows that each environment-aware refinement module independently enhances performance, with the combination yielding the best overall results.



\begin{table}
\centering
\resizebox{0.5\textwidth}{!}{
\begin{tabular}{l*{6}{c}}
\toprule
& \multicolumn{3}{c}{\textbf{EgoBody}} & \multicolumn{3}{c}{\textbf{GIMO}}  \\
\cmidrule(lr){2-4} \cmidrule(lr){5-7} 
& MPJRE & MPJPE & MPJVE & MPJRE & MPJPE & MPJVE \\
\midrule
1)Baseline w/o UNC & 7.63 & 90.8 & 333.0 & 5.90 & 77.5 & 270.2\\
2)Baseline & 7.45 & 87.1 & 342.9 & 5.72 & 74.6 & 266.7 \\
\midrule
3)EnvPoser w/o UNC & 6.56 & 80.1 & \textbf{172.7} & 4.45 & 60.2 & 278.6 \\
EnvPoser & \textbf{6.00} & \textbf{74.7} & 174.0 & \textbf{4.38} & \textbf{57.6} & \textbf{234.6}\\
\bottomrule
\end{tabular}
}
\vspace{-2mm}
\caption{The effectiveness of the uncertainty estimation.}
\vspace{-5mm}
\label{table:uncertainty}
\end{table}


\textbf{Effectiveness of the uncertainty estimation.}
As shown in Tab.~\ref{table:uncertainty}, we compare two variants: 1) Baseline w/o UNC: a baseline without uncertainty estimation and 2) Baseline: a baseline with uncertainty estimation. Results show that integrating the uncertainty estimation module not only quantifies joint uncertainty across motions but also improves full-body motion estimation through additional supervision. 
We further incorporated both baseline methods into the environment-aware refinement module. As shown in Tab.~\ref{table:uncertainty}, our method achieves significant improvements over variant 3) EnvPoser w/o UNC: our method without uncertainty estimation, reducing MPJRE and MPJPE errors by 8.11\% and 5.68\%, respectively, on the EgoBody dataset. These results indicate that the uncertainty estimation and resampling methods effectively capture the multi-hypothesis nature of initial motion estimates and, when combined with environmental refinement, adaptively align motion estimation with sparse tracking signals and environmental context. 






\section{Conclusions}
This paper presents a novel two-stage framework for reliable human motion estimation leveraging sparse tracking signals and environmental context. In the first stage, we model the multi-hypothesis nature of motion through uncertainty estimation, followed by a refinement phase that applies environment-aware constraints from both semantic and geometric perspectives. Extensive experiments show that EnvPoser outperforms existing methods, particularly in complex interaction scenarios, highlighting the significance of environmental context in wearable motion estimation and paving the way for future approaches to harness this valuable contextual information. This work has the potential to impact a wide range of interactive applications that benefit from more accurate and context-aware motion estimation.

\noindent{\bf Limitations and future works.} The model assumes a static environment, lacking consideration for dynamic multi-user interactions or object motion, which may limit its effectiveness. This limitation could be addressed by incorporating a third-person perspective to estimate the movement of other people and objects, along with additional semantic and geometric constraints. Furthermore, real-time mesh quality in complex environments is often limited, which may affect performance. Future work will explore using raw images to infer contacts in complex scenes via 2D semantic understanding and introducing additional constraints to improve overall reconstruction quality.



\clearpage
{
    \small
    \bibliographystyle{ieeenat_fullname}
    \bibliography{main}

\begin{thebibliography}{56}
\providecommand{\natexlab}[1]{#1}
\providecommand{\url}[1]{\texttt{#1}}
\expandafter\ifx\csname urlstyle\endcsname\relax
  \providecommand{\doi}[1]{doi: #1}\else
  \providecommand{\doi}{doi: \begingroup \urlstyle{rm}\Url}\fi

\bibitem[Aliakbarian et~al.(2022)Aliakbarian, Cameron, Bogo, Fitzgibbon, and
  Cashman]{aliakbarian2022flag}
Sadegh Aliakbarian, Pashmina Cameron, Federica Bogo, Andrew Fitzgibbon, and
  Thomas~J Cashman.
\newblock Flag: Flow-based 3d avatar generation from sparse observations.
\newblock In \emph{Proceedings of the IEEE/CVF Conference on Computer Vision
  and Pattern Recognition}, pages 13253--13262, 2022.

\bibitem[Aliakbarian et~al.(2023)Aliakbarian, Saleh, Collier, Cameron, and
  Cosker]{aliakbarian2023hmd}
Sadegh Aliakbarian, Fatemeh Saleh, David Collier, Pashmina Cameron, and Darren
  Cosker.
\newblock Hmd-nemo: Online 3d avatar motion generation from sparse
  observations.
\newblock In \emph{Proceedings of the IEEE/CVF International Conference on
  Computer Vision}, pages 9622--9631, 2023.

\bibitem[Chen et~al.(2023)Chen, Yang, and Yao]{chen2023mhentropy}
Rongyu Chen, Linlin Yang, and Angela Yao.
\newblock Mhentropy: Entropy meets multiple hypotheses for pose and shape
  recovery.
\newblock In \emph{Proceedings of the IEEE/CVF International Conference on
  Computer Vision}, pages 14840--14849, 2023.

\bibitem[Dai et~al.(2024)Dai, Zhang, Liu, Fan, Du, Su, Zheng, and
  Li]{dai2024hmd}
Peng Dai, Yang Zhang, Tao Liu, Zhen Fan, Tianyuan Du, Zhuo Su, Xiaozheng Zheng,
  and Zeming Li.
\newblock Hmd-poser: On-device real-time human motion tracking from scalable
  sparse observations.
\newblock In \emph{Proceedings of the IEEE/CVF Conference on Computer Vision
  and Pattern Recognition}, pages 874--884, 2024.

\bibitem[Dittadi et~al.(2021)Dittadi, Dziadzio, Cosker, Lundell, Cashman, and
  Shotton]{dittadi2021full}
Andrea Dittadi, Sebastian Dziadzio, Darren Cosker, Ben Lundell, Thomas~J
  Cashman, and Jamie Shotton.
\newblock Full-body motion from a single head-mounted device: Generating smpl
  poses from partial observations.
\newblock In \emph{Proceedings of the IEEE/CVF International Conference on
  Computer Vision}, pages 11687--11697, 2021.

\bibitem[Du et~al.(2023)Du, Kips, Pumarola, Starke, Thabet, and
  Sanakoyeu]{du2023avatars}
Yuming Du, Robin Kips, Albert Pumarola, Sebastian Starke, Ali Thabet, and
  Artsiom Sanakoyeu.
\newblock Avatars grow legs: Generating smooth human motion from sparse
  tracking inputs with diffusion model.
\newblock In \emph{Proceedings of the IEEE/CVF Conference on Computer Vision
  and Pattern Recognition}, pages 481--490, 2023.

\bibitem[Fan et~al.(2024)Fan, Dai, Su, Gao, Lv, Zhang, Du, Wang, and
  Zhang]{fan2024emhi}
Zhen Fan, Peng Dai, Zhuo Su, Xu Gao, Zheng Lv, Jiarui Zhang, Tianyuan Du,
  Guidong Wang, and Yang Zhang.
\newblock Emhi: A multimodal egocentric human motion dataset with hmd and
  body-worn imus.
\newblock \emph{arXiv preprint arXiv:2408.17168}, 2024.

\bibitem[Feng et~al.(2024)Feng, Ma, Gao, Zheng, Xue, and
  Xu]{feng2024stratified}
Han Feng, Wenchao Ma, Quankai Gao, Xianwei Zheng, Nan Xue, and Huijuan Xu.
\newblock Stratified avatar generation from sparse observations.
\newblock In \emph{Proceedings of the IEEE/CVF Conference on Computer Vision
  and Pattern Recognition}, pages 153--163, 2024.

\bibitem[Gong et~al.(2023)Gong, Foo, Fan, Ke, Rahmani, and
  Liu]{gong2023diffpose}
Jia Gong, Lin~Geng Foo, Zhipeng Fan, Qiuhong Ke, Hossein Rahmani, and Jun Liu.
\newblock Diffpose: Toward more reliable 3d pose estimation.
\newblock In \emph{Proceedings of the IEEE/CVF Conference on Computer Vision
  and Pattern Recognition}, pages 13041--13051, 2023.

\bibitem[Hassan et~al.(2019)Hassan, Choutas, Tzionas, and
  Black]{hassan2019resolving}
Mohamed Hassan, Vasileios Choutas, Dimitrios Tzionas, and Michael~J Black.
\newblock Resolving 3d human pose ambiguities with 3d scene constraints.
\newblock In \emph{Proceedings of the IEEE/CVF international conference on
  computer vision}, pages 2282--2292, 2019.

\bibitem[Huang et~al.(2022)Huang, Yi, H{\"o}schle, Safroshkin, Alexiadis,
  Polikovsky, Scharstein, and Black]{huang2022capturing}
Chun-Hao~P Huang, Hongwei Yi, Markus H{\"o}schle, Matvey Safroshkin, Tsvetelina
  Alexiadis, Senya Polikovsky, Daniel Scharstein, and Michael~J Black.
\newblock Capturing and inferring dense full-body human-scene contact.
\newblock In \emph{Proceedings of the IEEE/CVF Conference on Computer Vision
  and Pattern Recognition}, pages 13274--13285, 2022.

\bibitem[Huang et~al.(2023)Huang, Wang, Li, Jia, Liu, Zhu, Liang, and
  Zhu]{huang2023diffusion}
Siyuan Huang, Zan Wang, Puhao Li, Baoxiong Jia, Tengyu Liu, Yixin Zhu, Wei
  Liang, and Song-Chun Zhu.
\newblock Diffusion-based generation, optimization, and planning in 3d scenes.
\newblock In \emph{Proceedings of the IEEE/CVF Conference on Computer Vision
  and Pattern Recognition}, pages 16750--16761, 2023.

\bibitem[Huang et~al.(2018)Huang, Kaufmann, Aksan, Black, Hilliges, and
  Pons-Moll]{huang2018deep}
Yinghao Huang, Manuel Kaufmann, Emre Aksan, Michael~J Black, Otmar Hilliges,
  and Gerard Pons-Moll.
\newblock Deep inertial poser: learning to reconstruct human pose from sparse
  inertial measurements in real time.
\newblock \emph{ACM Trans. Graph. (TOG)}, 37\penalty0 (6):\penalty0 1--15,
  2018.

\bibitem[Jiang et~al.(2022{\natexlab{a}})Jiang, Streli, Qiu, Fender, Laich,
  Snape, and Holz]{jiang2022avatarposer}
Jiaxi Jiang, Paul Streli, Huajian Qiu, Andreas Fender, Larissa Laich, Patrick
  Snape, and Christian Holz.
\newblock Avatarposer: Articulated full-body pose tracking from sparse motion
  sensing.
\newblock In \emph{European conference on computer vision}, pages 443--460.
  Springer, 2022{\natexlab{a}}.

\bibitem[Jiang et~al.(2025)Jiang, Streli, Meier, and Holz]{jiang2025egoposer}
Jiaxi Jiang, Paul Streli, Manuel Meier, and Christian Holz.
\newblock Egoposer: Robust real-time egocentric pose estimation from sparse and
  intermittent observations everywhere.
\newblock In \emph{European Conference on Computer Vision}, pages 277--294.
  Springer, 2025.

\bibitem[Jiang et~al.(2024)Jiang, Zhang, Li, Ma, Wang, Chen, Liu, Zhu, and
  Huang]{jiang2024scaling}
Nan Jiang, Zhiyuan Zhang, Hongjie Li, Xiaoxuan Ma, Zan Wang, Yixin Chen, Tengyu
  Liu, Yixin Zhu, and Siyuan Huang.
\newblock Scaling up dynamic human-scene interaction modeling.
\newblock In \emph{Proceedings of the IEEE/CVF Conference on Computer Vision
  and Pattern Recognition}, pages 1737--1747, 2024.

\bibitem[Jiang et~al.(2022{\natexlab{b}})Jiang, Ye, Gopinath, Won, Winkler, and
  Liu]{jiang2022transformer}
Yifeng Jiang, Yuting Ye, Deepak Gopinath, Jungdam Won, Alexander~W Winkler, and
  C~Karen Liu.
\newblock Transformer inertial poser: Real-time human motion reconstruction
  from sparse imus with simultaneous terrain generation.
\newblock In \emph{SIGGRAPH Asia 2022 Conference Papers(SA' 22)}, pages 1--9,
  2022{\natexlab{b}}.

\bibitem[Kaufmann et~al.(2021)Kaufmann, Zhao, Tang, Tao, Twigg, Song, Wang, and
  Hilliges]{kaufmann2021pose}
Manuel Kaufmann, Yi Zhao, Chengcheng Tang, Lingling Tao, Christopher Twigg, Jie
  Song, Robert Wang, and Otmar Hilliges.
\newblock Em-pose: 3d human pose estimation from sparse electromagnetic
  trackers.
\newblock In \emph{Proceedings of the IEEE/CVF international conference on
  computer vision(ICCV)}, pages 11510--11520, 2021.

\bibitem[Kendall and Gal(2017)]{kendall2017uncertainties}
Alex Kendall and Yarin Gal.
\newblock What uncertainties do we need in bayesian deep learning for computer
  vision?
\newblock \emph{Advances in neural information processing systems}, 30, 2017.

\bibitem[Lee and Joo(2024)]{lee2024mocap}
Jiye Lee and Hanbyul Joo.
\newblock Mocap everyone everywhere: Lightweight motion capture with
  smartwatches and a head-mounted camera.
\newblock In \emph{Proceedings of the IEEE/CVF Conference on Computer Vision
  and Pattern Recognition}, pages 1091--1100, 2024.

\bibitem[Lee et~al.(2023)Lee, Starke, Ye, Won, and Winkler]{lee2023questenvsim}
Sunmin Lee, Sebastian Starke, Yuting Ye, Jungdam Won, and Alexander Winkler.
\newblock Questenvsim: Environment-aware simulated motion tracking from sparse
  sensors.
\newblock In \emph{ACM SIGGRAPH 2023 Conference Proceedings}, pages 1--9, 2023.

\bibitem[Li and Lee(2019)]{li2019generating}
Chen Li and Gim~Hee Lee.
\newblock Generating multiple hypotheses for 3d human pose estimation with
  mixture density network.
\newblock In \emph{Proceedings of the IEEE/CVF conference on computer vision
  and pattern recognition}, pages 9887--9895, 2019.

\bibitem[Li et~al.(2023)Li, Shi, Dai, Zheng, Wang, Sun, Guo, Li, Zou, and
  Xiong]{li2023pose}
Han Li, Bowen Shi, Wenrui Dai, Hongwei Zheng, Botao Wang, Yu Sun, Min Guo,
  Chenglin Li, Junni Zou, and Hongkai Xiong.
\newblock Pose-oriented transformer with uncertainty-guided refinement for
  2d-to-3d human pose estimation.
\newblock In \emph{Proceedings of the AAAI Conference on Artificial
  Intelligence}, pages 1296--1304, 2023.

\bibitem[Li et~al.(2022)Li, Liu, Tang, Wang, and Van~Gool]{li2022mhformer}
Wenhao Li, Hong Liu, Hao Tang, Pichao Wang, and Luc Van~Gool.
\newblock Mhformer: Multi-hypothesis transformer for 3d human pose estimation.
\newblock In \emph{Proceedings of the IEEE/CVF Conference on Computer Vision
  and Pattern Recognition}, pages 13147--13156, 2022.

\bibitem[Liu et~al.(2022)Liu, Hu, Lin, Yao, Xie, Wei, Ning, Cao, Zhang, Dong,
  et~al.]{liu2022swin}
Ze Liu, Han Hu, Yutong Lin, Zhuliang Yao, Zhenda Xie, Yixuan Wei, Jia Ning, Yue
  Cao, Zheng Zhang, Li Dong, et~al.
\newblock Swin transformer v2: Scaling up capacity and resolution.
\newblock In \emph{Proceedings of the IEEE/CVF conference on computer vision
  and pattern recognition}, pages 12009--12019, 2022.

\bibitem[Loper et~al.(2015)Loper, Mahmood, Romero, Pons-Moll, and
  Black]{loper2015smpl}
Matthew Loper, Naureen Mahmood, Javier Romero, Gerard Pons-Moll, and Michael~J
  Black.
\newblock Smpl: A skinned multi-person linear model.
\newblock \emph{ACM Transactions on Graphics}, 34\penalty0 (6), 2015.

\bibitem[Lou et~al.(2024)Lou, Cui, Wang, Tang, and Zhou]{lou2024multimodal}
Zhenyu Lou, Qiongjie Cui, Haofan Wang, Xu Tang, and Hong Zhou.
\newblock Multimodal sense-informed forecasting of 3d human motions.
\newblock In \emph{Proceedings of the IEEE/CVF Conference on Computer Vision
  and Pattern Recognition}, pages 2144--2154, 2024.

\bibitem[Mahmood et~al.(2019)Mahmood, Ghorbani, Troje, Pons-Moll, and
  Black]{mahmood2019amass}
Naureen Mahmood, Nima Ghorbani, Nikolaus~F Troje, Gerard Pons-Moll, and
  Michael~J Black.
\newblock Amass: Archive of motion capture as surface shapes.
\newblock In \emph{Proceedings of the IEEE/CVF international conference on
  computer vision}, pages 5442--5451, 2019.

\bibitem[Mihajlovic et~al.(2022)Mihajlovic, Saito, Bansal, Zollhoefer, and
  Tang]{mihajlovic2022coap}
Marko Mihajlovic, Shunsuke Saito, Aayush Bansal, Michael Zollhoefer, and Siyu
  Tang.
\newblock Coap: Compositional articulated occupancy of people.
\newblock In \emph{Proceedings of the IEEE/CVF Conference on Computer Vision
  and Pattern Recognition}, pages 13201--13210, 2022.

\bibitem[Ponton et~al.(2023)Ponton, Yun, Aristidou, Andujar, and
  Pelechano]{ponton2023sparseposer}
Jose~Luis Ponton, Haoran Yun, Andreas Aristidou, Carlos Andujar, and Nuria
  Pelechano.
\newblock Sparseposer: Real-time full-body motion reconstruction from sparse
  data.
\newblock \emph{ACM Transactions on Graphics}, 43\penalty0 (1):\penalty0 1--14,
  2023.

\bibitem[Ponton et~al.(2024)Ponton, Pujol, Aristidou, Andujar, and
  Pelechano]{ponton2024dragposer}
Jose~Luis Ponton, Eduard Pujol, Andreas Aristidou, Carlos Andujar, and Nuria
  Pelechano.
\newblock Dragposer: Motion reconstruction from variable sparse tracking
  signals via latent space optimization.
\newblock \emph{arXiv preprint arXiv:2406.14567}, 2024.

\bibitem[Qi et~al.(2017)Qi, Yi, Su, and Guibas]{qi2017pointnet++}
Charles~Ruizhongtai Qi, Li Yi, Hao Su, and Leonidas~J Guibas.
\newblock Pointnet++: Deep hierarchical feature learning on point sets in a
  metric space.
\newblock \emph{Advances in neural information processing systems}, 30, 2017.

\bibitem[Shan et~al.(2023)Shan, Liu, Zhang, Wang, Han, Wang, Ma, and
  Gao]{shan2023diffusion}
Wenkang Shan, Zhenhua Liu, Xinfeng Zhang, Zhao Wang, Kai Han, Shanshe Wang,
  Siwei Ma, and Wen Gao.
\newblock Diffusion-based 3d human pose estimation with multi-hypothesis
  aggregation.
\newblock In \emph{Proceedings of the IEEE/CVF International Conference on
  Computer Vision}, pages 14761--14771, 2023.

\bibitem[Shen et~al.(2023)Shen, Cen, Peng, Shuai, Bao, and
  Zhou]{shen2023learning}
Zehong Shen, Zhi Cen, Sida Peng, Qing Shuai, Hujun Bao, and Xiaowei Zhou.
\newblock Learning human mesh recovery in 3d scenes.
\newblock In \emph{Proceedings of the IEEE/CVF Conference on Computer Vision
  and Pattern Recognition}, pages 17038--17047, 2023.

\bibitem[Tang et~al.(2024)Tang, Wang, Ji, Xu, Yu, and Shi]{tang2024unified}
Jiangnan Tang, Jingya Wang, Kaiyang Ji, Lan Xu, Jingyi Yu, and Ye Shi.
\newblock A unified diffusion framework for scene-aware human motion estimation
  from sparse signals.
\newblock In \emph{Proceedings of the IEEE/CVF Conference on Computer Vision
  and Pattern Recognition}, pages 21251--21262, 2024.

\bibitem[Van~Wouwe et~al.(2024)Van~Wouwe, Lee, Falisse, Delp, and
  Liu]{van2024diffusionposer}
Tom Van~Wouwe, Seunghwan Lee, Antoine Falisse, Scott Delp, and C~Karen Liu.
\newblock Diffusionposer: Real-time human motion reconstruction from arbitrary
  sparse sensors using autoregressive diffusion.
\newblock In \emph{Proceedings of the IEEE/CVF Conference on Computer Vision
  and Pattern Recognition}, pages 2513--2523, 2024.

\bibitem[Vaswani(2017)]{vaswani2017attention}
A Vaswani.
\newblock Attention is all you need.
\newblock \emph{Advances in Neural Information Processing Systems}, 2017.

\bibitem[Wehrbein et~al.(2021)Wehrbein, Rudolph, Rosenhahn, and
  Wandt]{wehrbein2021probabilistic}
Tom Wehrbein, Marco Rudolph, Bodo Rosenhahn, and Bastian Wandt.
\newblock Probabilistic monocular 3d human pose estimation with normalizing
  flows.
\newblock In \emph{Proceedings of the IEEE/CVF international conference on
  computer vision}, pages 11199--11208, 2021.

\bibitem[Winkler et~al.(2022)Winkler, Won, and Ye]{winkler2022questsim}
Alexander Winkler, Jungdam Won, and Yuting Ye.
\newblock Questsim: Human motion tracking from sparse sensors with simulated
  avatars.
\newblock In \emph{SIGGRAPH Asia 2022 Conference Papers}, pages 1--8, 2022.

\bibitem[Wu et~al.(2024)Wu, chaoran wang, Yin, Guo, and Qin]{wu2024accurate}
Yinghao Wu, chaoran wang, Lu Yin, Shihui Guo, and Yipeng Qin.
\newblock Accurate and steady inertial pose estimation through sequence
  structure learning and modulation.
\newblock In \emph{The Thirty-eighth Annual Conference on Neural Information
  Processing Systems}, 2024.

\bibitem[Xia et~al.(2021)Xia, Chu, Pei, Zhang, Yu, and Qiu]{xia2021learning}
Songpengcheng Xia, Lei Chu, Ling Pei, Zixuan Zhang, Wenxian Yu, and Robert~C
  Qiu.
\newblock Learning disentangled representation for mixed-reality human activity
  recognition with a single imu sensor.
\newblock \emph{IEEE Transactions on Instrumentation and Measurement},
  70:\penalty0 1--14, 2021.

\bibitem[Xia et~al.(2024)Xia, Chu, Pei, Yang, Yu, and Qiu]{xia2024timestamp}
Songpengcheng Xia, Lei Chu, Ling Pei, Jiarui Yang, Wenxian Yu, and Robert~C
  Qiu.
\newblock Timestamp-supervised wearable-based activity segmentation and
  recognition with contrastive learning and order-preserving optimal transport.
\newblock \emph{IEEE Transactions on Mobile Computing}, 2024.

\bibitem[Xu et~al.(2024)Xu, Ma, Su, Zhu, Qiao, and Wang]{xu2024scorehypo}
Yuan Xu, Xiaoxuan Ma, Jiajun Su, Wentao Zhu, Yu Qiao, and Yizhou Wang.
\newblock Scorehypo: Probabilistic human mesh estimation with hypothesis
  scoring.
\newblock In \emph{Proceedings of the IEEE/CVF Conference on Computer Vision
  and Pattern Recognition}, pages 979--989, 2024.

\bibitem[Xuan et~al.(2023)Xuan, Li, Zhang, Zhang, Liu, and
  Li]{xuan2023narrator}
Haibiao Xuan, Xiongzheng Li, Jinsong Zhang, Hongwen Zhang, Yebin Liu, and Kun
  Li.
\newblock Narrator: Towards natural control of human-scene interaction
  generation via relationship reasoning.
\newblock In \emph{Proceedings of the IEEE/CVF International Conference on
  Computer Vision}, pages 22268--22278, 2023.

\bibitem[Yang et~al.(2024{\natexlab{a}})Yang, Li, Wu, Li, Wang, Yu, Su, and
  Xu]{yang2024smgdiff}
Hongdi Yang, Chengyang Li, Zhenxuan Wu, Gaozheng Li, Jingya Wang, Jingyi Yu,
  Zhuo Su, and Lan Xu.
\newblock Smgdiff: Soccer motion generation using diffusion probabilistic
  models.
\newblock \emph{arXiv preprint arXiv:2411.16216}, 2024{\natexlab{a}}.

\bibitem[Yang et~al.(2024{\natexlab{b}})Yang, Yao, and Ban]{yang2024spatial}
Xueyuan Yang, Chao Yao, and Xiaojuan Ban.
\newblock Spatial-related sensors matters: 3d human motion reconstruction
  assisted with textual semantics.
\newblock In \emph{Proceedings of the AAAI Conference on Artificial
  Intelligence}, pages 10225--10233, 2024{\natexlab{b}}.

\bibitem[Yi et~al.(2023)Yi, Huang, Tripathi, Hering, Thies, and
  Black]{yi2023mime}
Hongwei Yi, Chun-Hao~P Huang, Shashank Tripathi, Lea Hering, Justus Thies, and
  Michael~J Black.
\newblock Mime: Human-aware 3d scene generation.
\newblock In \emph{Proceedings of the IEEE/CVF Conference on Computer Vision
  and Pattern Recognition}, pages 12965--12976, 2023.

\bibitem[Yi et~al.(2021)Yi, Zhou, and Xu]{yi2021transpose}
Xinyu Yi, Yuxiao Zhou, and Feng Xu.
\newblock Transpose: Real-time 3d human translation and pose estimation with
  six inertial sensors.
\newblock \emph{ACM Transactions On Graphics (TOG)}, 40\penalty0 (4):\penalty0
  1--13, 2021.

\bibitem[Yi et~al.(2022)Yi, Zhou, Habermann, Shimada, Golyanik, Theobalt, and
  Xu]{yi2022physical}
Xinyu Yi, Yuxiao Zhou, Marc Habermann, Soshi Shimada, Vladislav Golyanik,
  Christian Theobalt, and Feng Xu.
\newblock Physical inertial poser (pip): Physics-aware real-time human motion
  tracking from sparse inertial sensors.
\newblock In \emph{Proceedings of the IEEE/CVF Conference on Computer Vision
  and Pattern Recognition(CVPR)}, pages 13167--13178, 2022.

\bibitem[Zhang et~al.(2021)Zhang, Zhang, Bogo, Pollefeys, and
  Tang]{zhang2021learning}
Siwei Zhang, Yan Zhang, Federica Bogo, Marc Pollefeys, and Siyu Tang.
\newblock Learning motion priors for 4d human body capture in 3d scenes.
\newblock In \emph{Proceedings of the IEEE/CVF International Conference on
  Computer Vision}, pages 11343--11353, 2021.

\bibitem[Zhang et~al.(2022)Zhang, Ma, Zhang, Qian, Kwon, Pollefeys, Bogo, and
  Tang]{zhang2022egobody}
Siwei Zhang, Qianli Ma, Yan Zhang, Zhiyin Qian, Taein Kwon, Marc Pollefeys,
  Federica Bogo, and Siyu Tang.
\newblock Egobody: Human body shape and motion of interacting people from
  head-mounted devices.
\newblock In \emph{European conference on computer vision}, pages 180--200.
  Springer, 2022.

\bibitem[Zhang et~al.(2023{\natexlab{a}})Zhang, Ma, Zhang, Aliakbarian, Cosker,
  and Tang]{zhang2023probabilistic}
Siwei Zhang, Qianli Ma, Yan Zhang, Sadegh Aliakbarian, Darren Cosker, and Siyu
  Tang.
\newblock Probabilistic human mesh recovery in 3d scenes from egocentric views.
\newblock In \emph{Proceedings of the IEEE/CVF International Conference on
  Computer Vision}, pages 7989--8000, 2023{\natexlab{a}}.

\bibitem[Zhang et~al.(2023{\natexlab{b}})Zhang, Wang, Kephart, and
  Ji]{zhang2023body}
Yufei Zhang, Hanjing Wang, Jeffrey~O Kephart, and Qiang Ji.
\newblock Body knowledge and uncertainty modeling for monocular 3d human body
  reconstruction.
\newblock In \emph{Proceedings of the IEEE/CVF International Conference on
  Computer Vision}, pages 9020--9032, 2023{\natexlab{b}}.

\bibitem[Zhang et~al.(2024)Zhang, Xia, Chu, Yang, Wu, and
  Pei]{zhang2024dynamic}
Yu Zhang, Songpengcheng Xia, Lei Chu, Jiarui Yang, Qi Wu, and Ling Pei.
\newblock Dynamic inertial poser (dynaip): Part-based motion dynamics learning
  for enhanced human pose estimation with sparse inertial sensors.
\newblock In \emph{Proceedings of the IEEE/CVF Conference on Computer Vision
  and Pattern Recognition}, pages 1889--1899, 2024.

\bibitem[Zheng et~al.(2023)Zheng, Su, Wen, Xue, and Jin]{zheng2023realistic}
Xiaozheng Zheng, Zhuo Su, Chao Wen, Zhou Xue, and Xiaojie Jin.
\newblock Realistic full-body tracking from sparse observations via joint-level
  modeling.
\newblock In \emph{Proceedings of the IEEE/CVF International Conference on
  Computer Vision}, pages 14678--14688, 2023.

\bibitem[Zheng et~al.(2022)Zheng, Yang, Mo, Li, Yu, Liu, Liu, and
  Guibas]{zheng2022gimo}
Yang Zheng, Yanchao Yang, Kaichun Mo, Jiaman Li, Tao Yu, Yebin Liu, C~Karen
  Liu, and Leonidas~J Guibas.
\newblock Gimo: Gaze-informed human motion prediction in context.
\newblock In \emph{European Conference on Computer Vision}, pages 676--694.
  Springer, 2022.

\end{thebibliography}
}


\clearpage
\setcounter{page}{1}
\setcounter{figure}{0}
\setcounter{table}{0}

\twocolumn[
	\begin{@twocolumnfalse}
		\section*{\centering{\Large{EnvPoser: Environment-aware Realistic Human Motion Estimation from Sparse Observations with Uncertainty Modeling}}}
		\begin{center}
			\large{Supplementary Material}


		\end{center}
		\centering
	\end{@twocolumnfalse}
]

\setcounter{section}{0}
\renewcommand\thesection{\Alph{section}}


This supplementary material provides additional information to complement our main paper. Sec.~\ref{sec:implementation} details the network architecture and training procedures of our proposed method. Sec.~\ref{sec:datasets} summarizes the datasets used in our study. Sec.~\ref{sec:add_analysis} presents extended ablation studies, encompassing various environment points sampling strategy and further analysis of component design. Sec.~\ref{sec:add_qualitative}  showcases additional qualitative comparisons between our approach and state-of-the-art methods. Finally, Sec.~\ref{sec:discussion} summarizes our work, discusses its limitations, and outlines potential directions for future research.

\section{Implementation Details}
\label{sec:implementation}
\textbf{Network Details.} Our model architecture consists of distinct sub-networks tailored to each stage. The uncertainty-aware initial motion estimation module processes sparse tracking signals and historical motion states, both pre-segmented into windows of length 40. These inputs are fed through two linear layers and positional encoding before entering a standard Transformer network with eight attention heads~\cite{vaswani2017attention}. Subsequently, two MLP layers estimate the human motion and corresponding joint uncertainty. During the first stage of training, this module is trained in two steps: first, the motion reconstruction head is optimized, followed by training the uncertainty estimation head using Eq.(3). The hyperparameters $\lambda_M$ and $\lambda_{\delta}$ are set to 1 and 0.001, respectively.

For environment-aware motion refinement module, we utilize PointNet++~\cite{qi2017pointnet++} to extract features from the cropped environment point clouds. These features are integrated into the motion-environment attention network. In this stage, the initial motion estimates, combined with additional inputs such as head translations and head height, are concatenated into a 175-dimensional motion embedding. This embedding is projected into a 256-dimensional latent space through a linear layer. Spatial priors are computed by normalizing scene points to a human-centered coordinate system and combining distance-based salience with directional components. These priors are refined using a learnable network with two fully connected layers and ReLU activations.
To integrate motion and environmental information, we employ a cross-attention mechanism that aligns scene features and spatial priors with the motion embedding. The resulting environment-refined motion representation $\boldsymbol{Z_{RM}}$ is obtained by passing the integrated features through an MLP with two fully connected layers.

Joint contact probabilities are predicted by projecting the concatenated environment-refined motion representation and sparse observations into a 256-dimensional embedding via a linear layer. This embedding is passed through a fully connected contact prediction head to estimate joint contacts.

Finally, the sparse observations and contact predictions are concatenated and processed through a decoder comprising two fully connected layers with ReLU activations, generating the final pose estimation.
During the second training phase, we use Eq.(12) as the objective function. At the beginning of this phase, the parameters of the first module are fixed, and only the environment refinement module is trained. The hyper-parameters ${\left\{ {\lambda_i} \right\}_{i=1,\dots,7}}$ are set to $\{ 2.0, 1.0, 0.75, 0.75, 0.75, 1.0, 0.1\}$.


\begin{figure}[t]
	\centering
	\includegraphics[width=8cm]{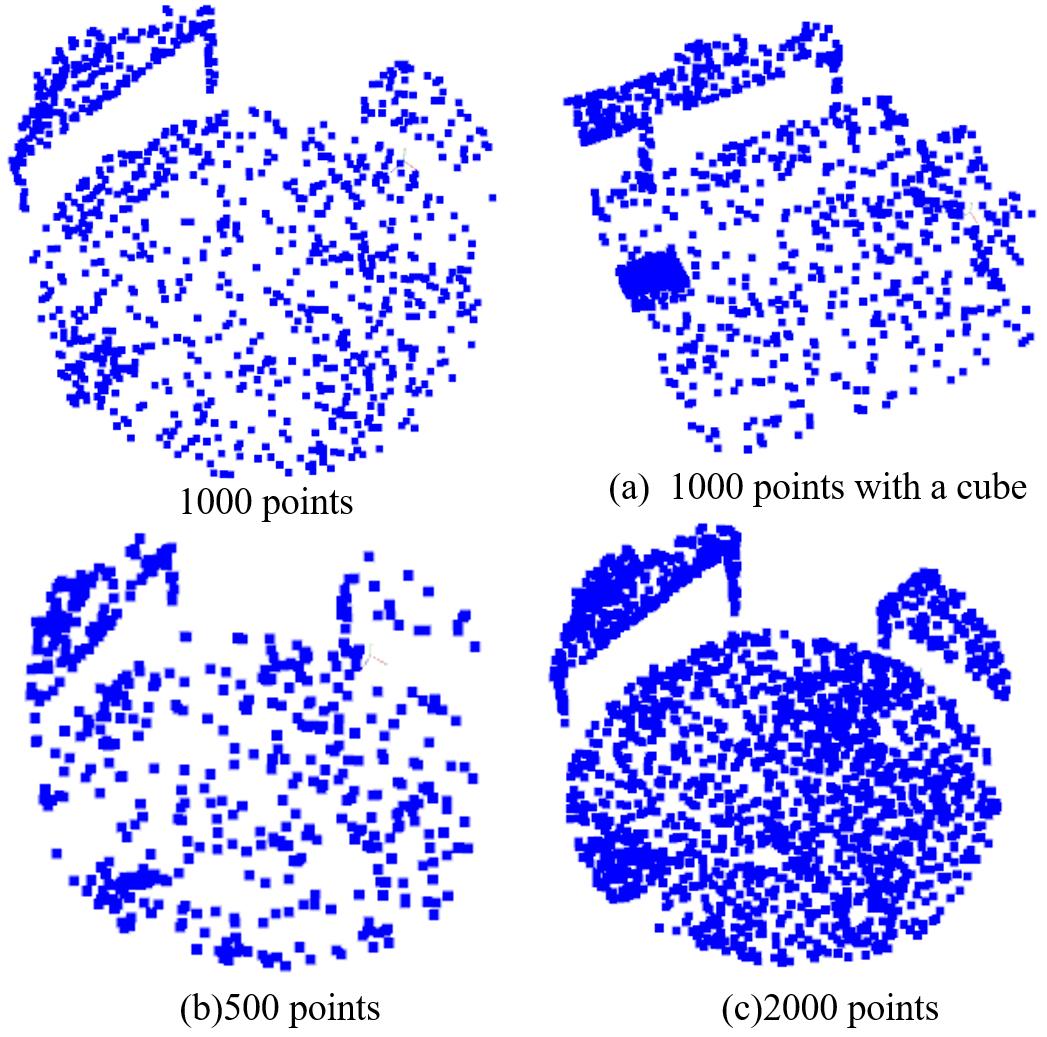}
	\caption{{Environmental point cloud with different sampling strategies.}}
	\label{environment_points}
	\vspace{-3mm}
\end{figure}

\begin{table}
\resizebox{0.45\textwidth}{!}{
\begin{tabular}{l*{4}{c}}
\toprule
& \multicolumn{2}{c}{\textbf{EgoBody}} & \multicolumn{2}{c}{\textbf{GIMO}}  \\
\cmidrule(lr){2-3} \cmidrule(lr){4-5} 
& MPJRE & MPJPE & MPJRE & MPJPE \\
\midrule
(a) EnvPoser-cube & 6.04 & 75.4 & 4.85 & 65.4 \\
(b) EnvPoser-500 & 6.05 & 76.4 & 4.47 & 60.2 \\
(c) EnvPoser-2000 & \textbf{5.95} & 76.0 & 4.45 & 59.8 \\  
\textbf{EnvPoser-1000} & 6.00 &\textbf{74.7} & \textbf{4.38} & \textbf{57.6} \\
\bottomrule
\end{tabular}
}
\caption{The effectiveness of environment point cloud sampling strategy.}
\label{table:environment_points}
\end{table}

\section{Datasets Details}
\label{sec:datasets}

To comprehensively evaluate the efficacy of our proposed method, we conduct experiments on two challenging public motion-scene interaction datasets: Egobody \cite{zhang2022egobody} and GIMO \cite{zheng2022gimo}. These datasets are carefully selected for their diverse and complex representations of human motion within interactive and immersive environments.

\begin{itemize}
    \item \textbf{Egobody:} The Egobody dataset is a robust egocentric dataset designed to capture 3D human motion during social interactions in immersive virtual environments. It includes 125 sequences collected from 36 participants, with an equal distribution of 18 males and 18 females. These participants engage in a wide range of social activities across 15 distinct indoor scenes, making the dataset highly diverse. Egobody offers detailed annotations for various interaction scenarios, enabling precise evaluation of motion estimation techniques in dynamic and interactive settings. Following the official split in \cite{zhang2022egobody}, the dataset is divided into 65 sequences for training and 43 sequences for testing. Its focus on immersive, egocentric perspectives provides valuable data for analyzing motion within constrained and socially active environments.


    \item \textbf{GIMO:} The GIMO dataset is notable for its multi-modal nature, offering a rich combination of body pose sequences, detailed environmental scans, and eye gaze data. Motion data were collected using a combination of HoloLens devices and IMU-based motion capture suits, providing precise motion trajectories and body pose sequences. Additionally, an iPhone 12 was used to scan the surrounding ambient scenes, resulting in high-quality environmental reconstructions. This multi-modal setup enables the dataset to capture the intricate relationships between human motion and environmental contexts. GIMO is particularly valuable for its emphasis on interactions within detailed 3D environments, offering a comprehensive perspective that bridges the gap between motion and scene understanding.
\end{itemize}

To ensure a diverse range of motion data, we also leverage the AMASS~\cite{mahmood2019amass} dataset during training. Specifically, our model's uncertainty-aware initial human motion estimation module is first trained on AMASS to capture diverse motion patterns. Subsequently, the entire model is fine-tuned on motion-environment interaction datasets~\cite{zhang2022egobody, zheng2022gimo}, incorporating the environment-aware motion refinement module in the second stage.




\begin{table}
\resizebox{0.45\textwidth}{!}{
\begin{tabular}{l*{4}{c}}
\toprule
& \multicolumn{2}{c}{\textbf{EgoBody}} & \multicolumn{2}{c}{\textbf{GIMO}}  \\
\cmidrule(lr){2-3} \cmidrule(lr){4-5} 
& MPJRE & MPJPE & MPJRE & MPJPE \\
\midrule
EnvPoser-w/o Contact & 6.18 & 77.6 & 4.49 & 59.4 \\  
\textbf{EnvPoser} & \textbf{6.00} &\textbf{74.7} & \textbf{4.38} & \textbf{57.6} \\
\bottomrule
\end{tabular}
}
\vspace{-2mm}
\caption{The effectiveness of contact estimation module.}
\label{table:contact}
\vspace{-3mm}
\end{table}

\begin{figure}[t]
	\centering
	\includegraphics[width=8cm]{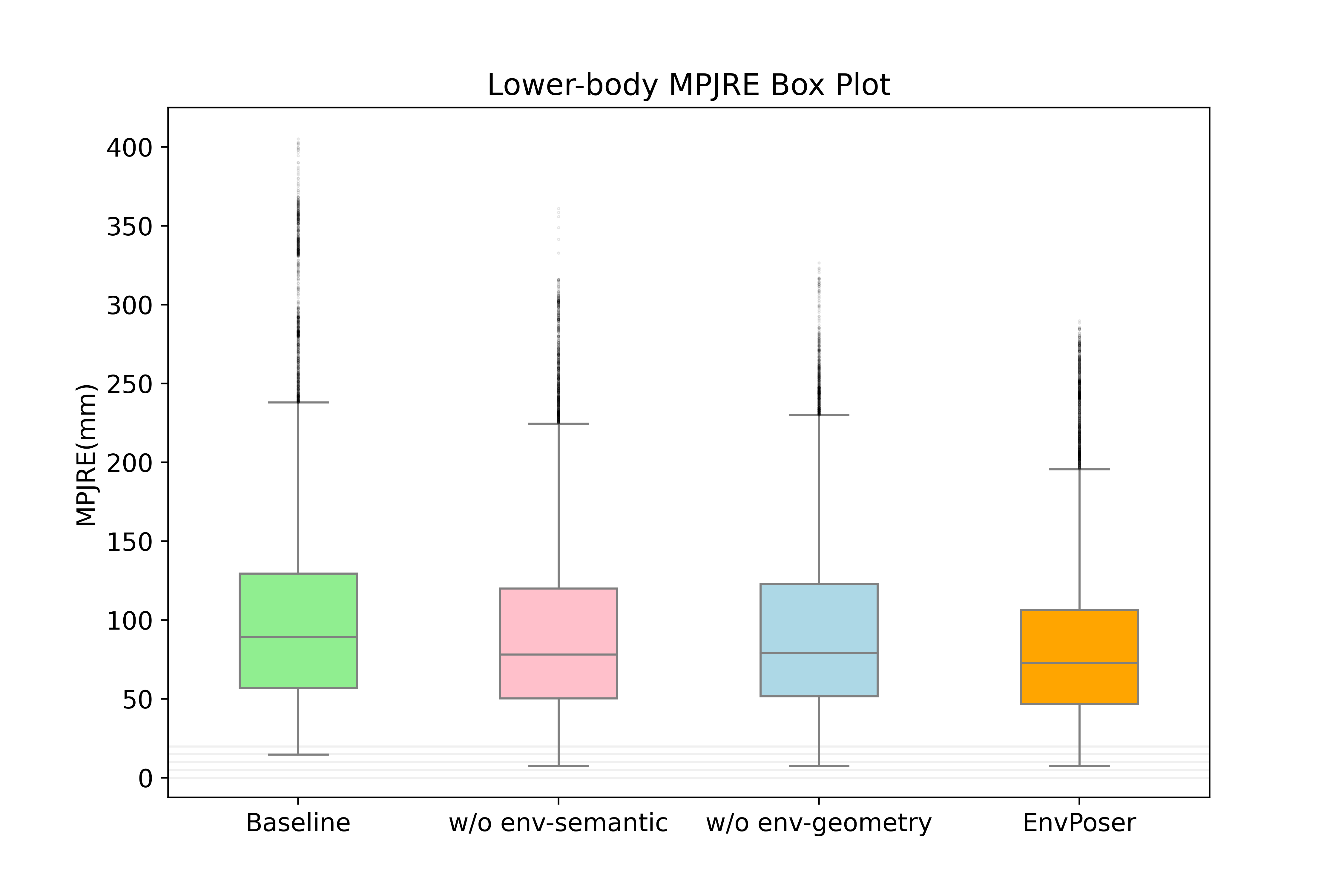}
	\caption{{Qualitative results of lower-body MPJPE box plot for ablation study on GIMO dataset.}}
	\label{boxplot}
\end{figure}

\begin{figure*}[!t]
\centering
	\includegraphics[width=17cm]{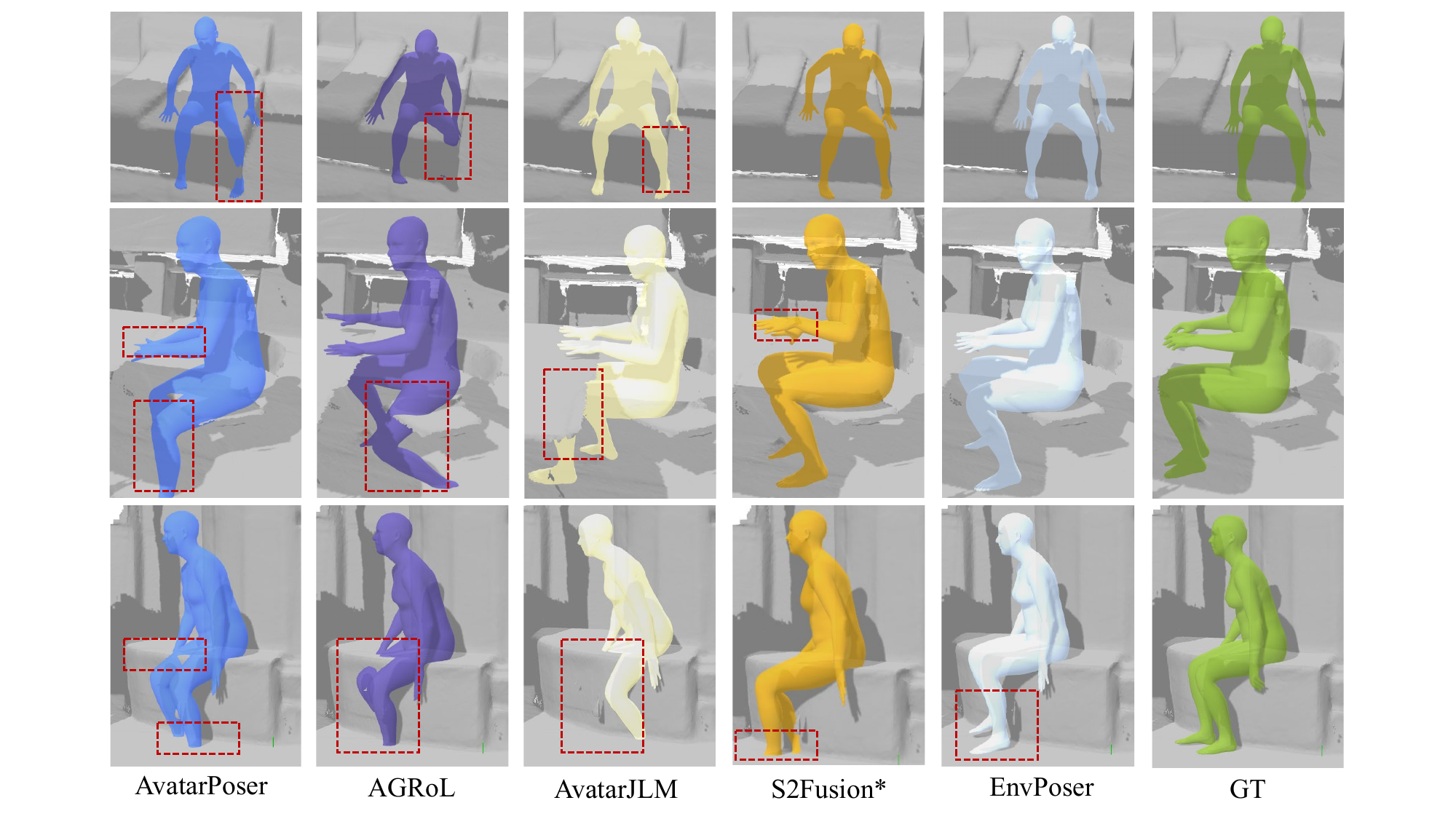}
    \vspace{-2mm}
	\caption{Visualization of various sitting motions from EgoBody and GIMO Datasets.}
	\label{fig:cvpr25_vis_sitting}
    \vspace{-3mm}
\end{figure*}

\section{Additional Analysis}
\label{sec:add_analysis}

In this section, we present more experimental results of our proposed method (EnvPoser).

 
\textbf{{Effectiveness of the Environment Point Clouds Sampling Strategy.}} EnvPoser processes the environmental point cloud through clipping and normalization in the Environmental Point Cloud Embedding module. Specifically, we first clip the environmental point clouds based on the human’s global position, as illustrated in Fig.~\ref{environment_points}. To evaluate the impact of the clipping and sampling strategies on model performance, we experimented with various sampling methods. In this study, we replaced the circular sampling approach with a square sampling method, shown in Fig.~\ref{environment_points}(a). In Tab.~\ref{table:environment_points}, we could find that using circular sampling improves motion estimation performance to some extent.

Subsequently, based on circular clipping, we adjusted the number of sampled points to 500 and 2000 to further analyze the impact of point cloud sampling strategies on model performance. 
Sampling 500 environmental points as input leads to performance degradation due to the sparsity of the point cloud. Conversely, increasing the number of sampled points to 2000 does not consistently improve performance. On the EgoBody dataset, EnvPoser-2000 achieves a marginal improvement of 0.05 in the MPJRE metric, but on the GIMO dataset, using more points results in a decline in performance. These results indicate that sampling 1000 environmental points within the selected area is sufficient to refine the initial motion estimates. Considering the trade-off between motion estimation accuracy and computational efficiency, this paper adopts circular sampling with 1000 environmental points as the optimal configuration.

\textbf{{Effectiveness of the Contact Estimation Module.}}
With the environment-refined motion representation $\boldsymbol{Z}_{RM}$ obtained through environment-semantic attention, EnvPoser first estimates contact probabilities and subsequently regresses full-body motion. To validate the effectiveness of the contact probability estimation, we conducted an ablation experiment by removing this step and directly regressing full-body motion from the environment-refined motion representation $\boldsymbol{Z}_{RM}$. This variant, referred to as EnvPoser-w/o Contact, was evaluated on the EgoBody and GIMO datasets.

As shown in Tab.~\ref{table:contact}, omitting the contact probability estimation results in degraded motion reconstruction performance compared to EnvPoser. Specifically, the MPJPE metric declines by 3.7\% on the EgoBody dataset and 3.0\% on the GIMO dataset, highlighting the importance of contact probability estimation in enhancing reconstruction accuracy.


\textbf{{Additional Comparison on Environment Refinement Module.}} To further illustrate the effectiveness of our environment refinement module, we present box plots of lower-body position estimation errors (Lower-body MPJPE) on the Gimo dataset for EnvPoser and its three variants: \ding{172} Baseline, \ding{173} w/o env-semantic, and \ding{174} w/o env-geometry, as defined in Sec.4.

As shown in Fig.~\ref{boxplot}, EnvPoser achieves the lowest maximum Lower-body MPJPE and exhibits significantly fewer outliers, demonstrating that the integration of semantic and geometric environmental constraints enables the most robust and accurate estimation across diverse motions. Comparing variants \ding{173} and \ding{174} with \ding{172}, we observe that incorporating environmental information significantly improves lower-body motion estimation accuracy. The constraints provided by environmental information not only enhance overall accuracy but also effectively reduce outliers in lower-body estimation, underscoring the importance of leveraging both environmental semantic and geometric aspects for refining motion estimates.

\begin{figure*}[!t]
\centering
	\includegraphics[width=17cm]{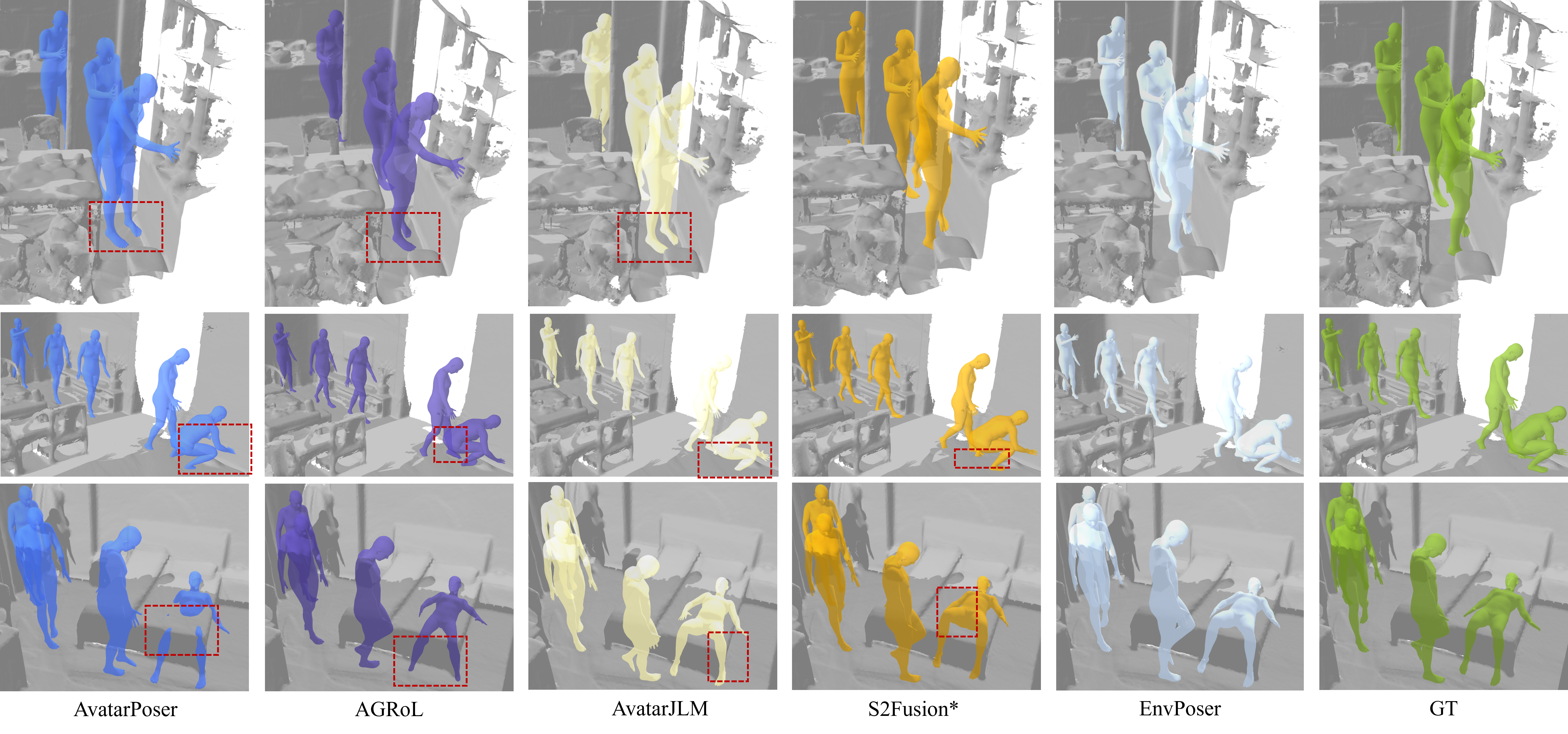}
    \vspace{-2mm}
	\caption{Visualization of full-body estimation on three test sequences from GIMO Datasets.}
	\label{fig:cvpr25_gimo}
    \vspace{-3mm}
\end{figure*}

\section{Additional Qualitative Results}
\label{sec:add_qualitative}
In this section, We show more visualization results of our method compared to the state-of-the-art methods.

\textbf{Additional Qualitative Comparison on Sitting Motion.} The most common interaction between human motion and the environment involves sitting, such as sitting on chairs, sofas, or beds. Fig.~\ref{fig:cvpr25_vis_sitting} compares the performance of our method, EnvPoser, with other approaches in visualizing sitting motions across various scenarios. Notably, EnvPoser effectively adapts to different object shapes, generating realistic and plausible sitting poses. In contrast, methods that lack environmental information~\cite{zheng2023realistic}, such as AvatarPoser~\cite{jiang2022avatarposer} and AGRoL~\cite{du2023avatars}, can estimate sitting motions but often produce diverse and inconsistent lower-body poses due to the absence of joint observations. This results in unrealistic and impractical sitting motion estimates that fail to align with actual human motions.

S2Fusion~\cite{tang2024unified}, which incorporates environmental information, shows notable improvements in sitting motion estimation compared to other competing methods. However, it performs slightly less accurately than EnvPoser in estimating upper-body interactions with objects during sitting motions. These findings highlight the effectiveness of EnvPoser, which leverages environmental information from both semantic and geometric perspectives to refine full-body motion estimation, ensuring more accurate and contextually appropriate results.

\textbf{Additional Qualitative Comparison on GIMO Dataset.}
Fig.~\ref{fig:cvpr25_gimo} presents motion estimation results on the GIMO dataset, showcasing three representative sequences with human motion visualized at key moments. Errors in comparison methods are highlighted with red bounding boxes for clarity. As illustrated in Fig.~\ref{fig:cvpr25_gimo}, EnvPoser exhibits robust performance, effectively handling challenging scenarios such as lying down and navigating narrow passages. These results further demonstrate the effectiveness and reliability of our proposed method.

\begin{figure}[t]
	\centering
	\includegraphics[width=8cm]{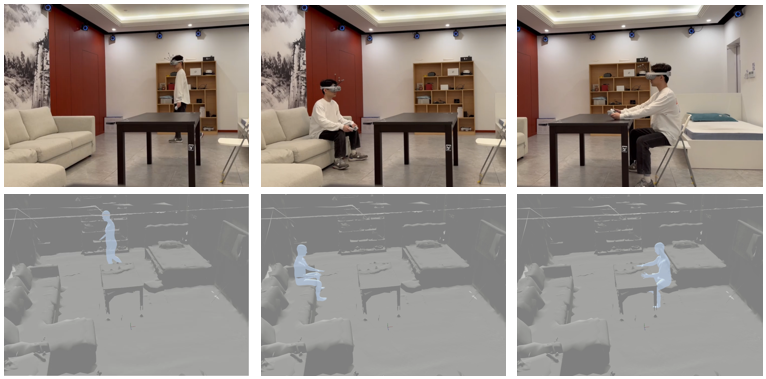}
	\caption{{Qualitative results on real data from VR device.}}
	\label{realmocap}
	\vspace{-3mm}
\end{figure}

\textbf{Additional Qualitative Comparison on Realistic Self-collected Data.}
We validated our model using data collected from real-world VR devices. The environment mesh within the motion range was pre-scanned, and human motion was estimated based on three 6DoF tracking signals captured by the VR devices. Fig.~\ref{realmocap} showcases the full-body motion estimation results of EnvPoser, with a more comprehensive performance demonstration available in the supplementary video material.

\section{Discussions and Future Works}
\label{sec:discussion}
Despite the effectiveness of our proposed framework, there are limitations that provide opportunities for future research and development.

\textbf{Static Environment Assumptions: }
Our current model assumes a static environment, which does not account for dynamic interactions involving multiple users or moving objects. While this simplification facilitates efficient motion estimation, it limits the model's applicability in real-world scenarios where dynamic changes are common. For example, interactions in crowded spaces or with moving objects, such as passing a ball, are not effectively captured. To address this, future work could incorporate a third-person perspective or additional external sensors to estimate the movements of other users and objects. These additions could complement our existing environmental refinement module by providing more robust motion-context interaction capabilities.

\textbf{Mesh Quality in Complex Environments: }
In real-time applications, the quality of pre-scanned environment meshes can vary significantly, especially in complex environments. Low-quality meshes may introduce inaccuracies in environmental constraints, impacting overall motion estimation performance. Exploring methods to enhance real-time mesh quality or mitigate the effects of noisy environment inputs will be essential.

\textbf{Leveraging Raw Ego-centric Visual Data: }
Lastly, the model currently relies on pre-scanned point clouds for environmental context. Future work could extend this by incorporating raw visual data, such as images or video streams, to infer contact points and environmental semantics directly through 2D or 3D understanding. This approach could enable real-time processing of dynamic environments while reducing the dependency on pre-scanned data. For example, 2D semantic segmentation combined with depth estimation could enhance the system's ability to handle occlusions and complex interactions in cluttered scenes.

In summary, while our method achieves state-of-the-art performance in human motion estimation with sparse tracking signals, addressing these limitations through dynamic interactions, improved environmental modeling, and adaptive strategies will further enhance its robustness and applicability. These directions hold promise for broadening the utility of our approach in increasingly complex and interactive environments.


\end{document}